\documentclass{article}

\usepackage{arxiv}

\usepackage[utf8]{inputenc} 
\usepackage[T1]{fontenc}    
\usepackage{hyperref}       
\usepackage{url}            
\usepackage{booktabs}       
\usepackage{amsfonts}       
\usepackage{nicefrac}       
\usepackage{microtype}      
\usepackage{lipsum}		
\usepackage{microtype}
\usepackage{graphicx}
\usepackage{subfigure}
\usepackage{booktabs} 
\usepackage{amsmath}
\usepackage{amssymb}
\usepackage{natbib}

\usepackage{hyperref}
\usepackage{bm}
\usepackage{xr}
\usepackage{diagbox}
\RequirePackage{fancyhdr}
\RequirePackage{color}
\RequirePackage{algorithm}
\RequirePackage{algorithmic}
\RequirePackage{natbib}
\RequirePackage{eso-pic} 
\RequirePackage{forloop}
\newcommand{\Mean}{{\mathbb{E}}}
\newcommand{\Var}{{\mbox{Var}}}
\newcommand{\Cov}{{\mbox{cov}}}

\newcommand{\prob}{{\mathbb{P}}}
\DeclareMathOperator*{\argmin}{arg\,min}

\newtheorem{thm}{Theorem}
\newtheorem{lemma}{Lemma}

\graphicspath{{figs0/}}

\makeatletter
\newcommand{\printfnsymbol}[1]{%
	\textsuperscript{\@fnsymbol{#1}}%
}
\makeatother

\title{Deeply-Debiased Off-Policy Interval Estimation}

\author{
	Chengchun Shi\thanks{Equal contribution}\\
	London School of Economics and Political Science
   \And
  Runzhe Wan\printfnsymbol{1} \\
  North Carolina State University
  \AND
  Victor Chernozhukov \\
  Massachusetts Institute of Technology
  \And
  Rui Song\\
  North Carolina State University
}
\date{}

\begin{document}
\maketitle

\begin{abstract}
	Off-policy evaluation learns a target policy's value with a historical dataset generated by a different behavior policy. In addition to a point estimate, many applications would benefit significantly from having a confidence interval (CI) that quantifies the uncertainty of the point estimate. 
	In this paper, we propose a novel deeply-debiasing procedure to construct an efficient, robust, and flexible CI on a target policy's value. 
	Our method is justified by theoretical results and numerical experiments. A \textsf{Python} implementation
	of the proposed procedure is available at \url{https://github.com/RunzheStat/D2OPE}.
\end{abstract}


\section{Introduction}
\label{secintroduction}

Reinforcement learning \citep[RL,][]{sutton2018reinforcement} is a general technique in sequential decision making that learns an optimal policy to maximize the average cumulative reward. Prior to adopting any policy in practice, it is crucial to know the impact of implementing such a policy. In many real domains such as healthcare \citep{murphy2001marginal,luedtke2017evaluating,shi2020breaking}, robotics \citep{andrychowicz2020learning} and autonomous driving \citep{sallab2017deep}, it is costly, risky, unethical, or even infeasible to evaluate a policy's impact by directly running this policy. 
This motivates us to study the off-policy evaluation (OPE) problem that learns a target policy's value with pre-collected data generated by a different behavior policy. 

In many applications (e.g., mobile health studies), the number of observations is limited. Take the OhioT1DM dataset \citep{marling2018ohiot1dm} as an example, only a few thousands observations are available \citep{shi2020does}. In these cases, in addition to a point estimate on a target policy's value, it is crucial to construct a confidence interval (CI) that quantifies the uncertainty of the value estimates. 



This paper is concerned with the following question: 
\textit{is it possible to develop a robust and efficient off-policy value estimator, and provide rigorous uncertainty quantification under practically feasible conditions?} 
We will give an affirmative answer to this question. 

\textbf{Overview of the OPE Literature}. There is a growing literature for OPE. Existing works can be casted into as direct method \citep[see e.g.,][]{le2019batch,shi2020statistical,feng2020accountable}, importance sampling-based method \citep[IS,][]{precup2000eligibility,thomas2015high2,hanna2016bootstrapping,liu2018breaking,nachum2019dualdice,dai2020coindice} and doubly robust method \citep{jiang2016doubly,thomas2016data,farajtabar2018more,tang2019doubly,uehara2019minimax,kallus2020double,jiang2020minimax}. Direct method derives the value estimates by learning the system transition matrix or the Q-function under the target policy. IS estimates the value by re-weighting the observed rewards with the density ratio of the target and behavior policies. Both direct method and IS have their own merits. In general, IS-type estimators might suffer from a large variance due to the use of the density ratio, whereas direct method might suffer from a large bias due to the potential misspecification of the model. Doubly robust methods combine both for more robust and efficient value evaluation.

Despite the popularity of developing a point estimate of a target policy's value, less attention has been paid to constructing its CI, which is the focus of this paper. Among those available, \citet{thomas2015high2} and \citet{hanna2016bootstrapping} derived the CI by using bootstrap or concentration inequality applied to the stepwise IS estimator. These methods suffer from the curse of horizon \citep{liu2018breaking}, leading to very large CIs. \citet{feng2020accountable} applied the Hoeffding's inequality to derive the CI based on a kernel-based Q-function estimator. Similar to the direct method, their estimator might suffer from a large bias. 
\citet{dai2020coindice} reformulated the OPE problem using the generalized estimating equation approach and applied the empirical likelihood approach \citep[see e.g.,][]{owen2001empirical} to CI estimation. They derived the CI by assuming the data transactions are i.i.d. However, observations in reinforcement learning are typically time-dependent. Directly applying the empirical likelihood method to weakly dependent data would fail without further adjustment \citep{kitamura1997empirical,duchi2016statistics}. The resulting CI might not be valid. We discuss this in detail in Appendix \ref{sec:coindice}.

Recently, \citet{kallus2019efficiently} made an important step forward for OPE, by developing a double reinforcement learning (DRL) estimator that achieves the semiparametric efficiency bound \citep[see e.g.,][]{tsiatis2007semiparametric}. Their method learns a Q-function and a marginalized density ratio and requires either one of the two estimators to be consistent. When both estimators converge at certain rates, DRL is asymptotically normal, based on which a Wald-type CI can be derived. 
However, these convergence rates might not be achievable in complicated RL tasks with high-dimensional state variables, resulting in an asymptotically biased value estimator and an invalid CI. See Section \ref{secDROPE} for details.

Finally, we remark that our work is also related to a line of research on statistical inference in bandits \citep{van2014online,deshpande2018accurate,zhang2020inference,hadad2021confidence}. However, these methods are not applicable to our setting.

\textbf{Advances of the Proposed Method.} Our proposal is built upon the DRL estimator to achieve sample efficiency. To derive a valid CI under weaker and practically more feasible  conditions than DRL, we propose to learn a conditional density ratio estimator and develop a  deeply-debiasing process that iteratively reduces the biases of the Q-function and value estimator. Debiasing brings additional robustness and flexibility. In a contextual bandit setting, our proposal shares similar spirits to the minimax optimal estimating procedure that uses higher order influence functions for learning the average treatment effects \citep[see e.g.,][]{robins2008higher,robins2017minimax,mukherjee2017semiparametric,mackey2018orthogonal}.
As such, the proposed method is: 
\begin{itemize}
	\item \textbf{robust} as the proposed value estimator is more robust than DRL and can converge to the true value in cases where neither the Q-function nor the marginalized density ratio estimator is consistent. More specifically, it is ``triply-robust" and requires the Q-function,  marginalized density ratio, or conditional density ratio estimator to be consistent. See Theorem \ref{thm_TR} for a formal statement. 
	
	\item \textbf{efficient} as we can show it achieves the semiparametric efficiency bound as DRL. 
	This in turn implies that the proposed CI is tight. See Theorem \ref{thm_eff} for details. 
	
	
	\item \textbf{flexible} as it requires much weaker and practically more feasible conditions to achieve nominal coverage. Specifically, our procedure allows the Q-estimator and marginalized density ratio to converge at an arbitrary rate. See Theorem \ref{thm_adaptive} for details. 
\end{itemize}

\section{Preliminaries}
\label{secpreliminaries}
We first formulate the OPE problem. We next review the DRL method, as it is closely related to our proposal. 
\subsection{Off-Policy Evaluation}
\label{secOPE}

We assume the data in OPE follows a 
Markov Decision Process \citep[MDP,][]{puterman2014markov} model defined by a tuple 
$(\mathcal{S}, \mathcal{A}, p, r, \gamma)$, 
where $\mathcal{S}$ is the state space, $\mathcal{A}$ is the action space, 
$p: \mathcal{S}^2 \times \mathcal{A}  \rightarrow [0,1]$ is the Markov transition matrix that characterizes the system transitions, $r : \mathcal{S} \times \mathcal{A} \rightarrow \mathbb{R}$ is the reward function, 
and $\gamma \in (0, 1)$ is a discounted factor that balances the immediate and future rewards. To simplify the presentation, we assume the state space is discrete. Meanwhile, the proposed method is equally applicable to continuous state space as well. 

Let $\{(S_t,A_t,R_t)\}_{t\ge 0}$ denote a trajectory generated from the MDP model where $(S_t,A_t,R_t)$ denotes the state-action-reward triplet at time $t$. 
Throughout this paper, we assume the following  Markov assumption (MA) and the conditional mean independence assumption (CMIA) hold: 
\begin{align*}
&\prob(S_{t+1}=s|\{S_{j},A_{j},R_{j}\}_{0\le j\le t})=p(s;A_{t},S_{t}),\,\,(\hbox{MA}),\\ 
&\Mean(R_{t}|S_{t},A_{t},\{S_{j},A_{j},R_{j}\}_{0\le j< t})=r(A_{t},S_{t}),\,\,(\hbox{CMIA}).
\end{align*}
These two assumptions guarantee the existence of an optimal \textit{stationary} policy \citep[see e.g.,][]{puterman2014markov}. Following a given stationary policy $\pi$, the agent will select action $a$ with probability $\pi(a|s)$ at each decision time. 
The corresponding state value function and state-action value function (better known as the Q-function) are given as follows:
\begin{align*}
V^{\pi}(s)&=\sum_{t=0}^{+\infty} \gamma^t\Mean^{\pi} ( R_{t}|S_{0}=s),\\
Q^{\pi}(a,s)&= \sum_{t=0}^{+\infty} \gamma^t \Mean^{\pi}(R_{t}|A_{0}=a,S_{0}=s),
\end{align*}
where the expectation $\Mean^{\pi}$ is defined by assuming the system follows the policy $\pi$. 

The observed data consists of $n$ i.i.d. trajectories, and can be summarized as $\{(S_{i,t},A_{i,t},R_{i,t},S_{i,t+1})\}_{0\le t<T_i,1\le i\le n}$ where $T_i$ denotes the termination time of the $i$th trajectory. Without loss of generality, we assume $T_1=\cdots=T_n=T$ and the immediate rewards are uniformly bounded. 
We consider evaluating the value of a given \textit{target policy} $\pi$ with respect to a given reference distribution $\mathbb{G}$, defined as 
\begin{eqnarray*}
	\eta^{\pi} =  \Mean_{s \sim \mathbb{G}} V^{\pi}(s).
\end{eqnarray*} 
In applications such as video games where a large number of trajectories are available, one may set $\mathbb{G}$ to the initial state distribution and approximate it by the empirical distribution of $\{S_{i,0}\}_{1\le i\le n}$. In applications such as mobile health studies, the number of trajectories is limited. For instance, the OhioT1DM dataset contains data for six patients (trajectories) only. In these cases, $\mathbb{G}$ shall be manually specified. In this paper, we primarily focus on the latter case with a prespecified $\mathbb{G}$. Meanwhile, the proposed method is equally applicable to the former case as well.
%
%


\subsection{Double Reinforcement Learning}
\label{secDROPE}
We review the DRL estimator in this section. 
We first define the marginalized density ratio under the target policy $\pi$ as  
\begin{eqnarray}\label{eqn:omega}
\omega^{\pi}(a,s)=\frac{(1-\gamma)\sum_{t=0}^{+\infty} \gamma^{t} p_t^{\pi}(a,s)}{p_{\infty}(a,s)},
\end{eqnarray}
where $p_t^{\pi}(a,s)$ denotes the probability of $(A_t,S_t)=(a,s)$ following policy $\pi$ with $S_{0}\sim \mathbb{G}$, and $p_{\infty}$ denotes the limiting distribution of the stochastic process $\{(A_t,S_t)\}_{t\ge 0}$. 
Such a marginalized density ratio plays a critical role in 
breaking the curse of horizon. 


Let $\widehat{Q}$ and $\widehat{\omega}$ be some estimates for $Q^{\pi}$ and $\omega^{\pi}$, respectively. 
\citet{kallus2019efficiently} proposed to construct the following estimating function for every $i$ and $t$: 
\begin{eqnarray}\label{term}
\begin{split}
\psi_{i,t}
\equiv
\frac{1}{1-\gamma}\widehat{\omega}(A_{i,t},S_{i,t})\{R_{i,t} 
-\widehat{Q}(A_{i,t},S_{i,t})+
\gamma 
\Mean_{a \sim \pi(\cdot| S_{i,t+1})}\widehat{Q}(a, S_{i,t+1})\}
+ \Mean_{s \sim \mathbb{G}, a \sim \pi(\cdot| s)}\widehat{Q}(a, s). 
\end{split}	
\end{eqnarray}
The resulting value estimator is given by
\begin{eqnarray*}
	\widehat{\eta}_{\tiny{\textrm{DRL}}}=\frac{1}{nT}\sum_{i=1}^n\sum_{t=0}^{T-1} \psi_{i,t}.
\end{eqnarray*}
One can show that 
$\widehat{\eta}_{\tiny{\textrm{DRL}}}$ is consistent when either $\widehat{Q}$ or $\widehat{\omega}$ is consistent. This is referred to as the doubly-robustness property. 
In addition, 
when both $\widehat{Q}$ and $\widehat{\omega}$ converge at a rate faster than $(nT)^{-1/4}$, 
$\sqrt{nT} (\widehat{\eta}_{\tiny{\textrm{DRL}}} - \eta^{\pi})$ converges weakly to a normal distribution with mean zero and variance
\begin{eqnarray}\label{lower_bound}
\frac{1}{(1-\gamma)^2}\Mean \left[ 
\omega^{\pi}(A, S) \{R + \gamma V^{\pi}(S') -  Q^{\pi}(A,S)\}
\right]^2,
\end{eqnarray}
where the tuple $(S,A,R,S')$ follows the limiting distribution of the process $\{(S_t,A_t,R_t,S_{t+1})\}_{t\ge 0}$. See Theorem 11 of \citet{kallus2019efficiently} for a formal proof. 
A consistent estimator for \eqref{lower_bound} can be derived based on the observed data. A Wald-type CI for $\eta^{\pi}$ can thus be constructed. 

Moreover, it follows from Theorem 5 of \citet{kallus2019efficiently} that \eqref{lower_bound} is 
the \textit{semiparametric efficiency bound} for infinite-horizon OPE. Informally speaking,  a semiparametric efficiency bound can be viewed as the nonparametric extension of the Cramer–Rao lower bound in parametric models \cite{bickel1993efficient}. It provides a lower bound of the asymptotic variance among all regular estimators \cite{van2000asymptotic}. 
Many other OPE methods such as \citet{liu2018breaking}, are statistically inefficient in that the variance of their value estimator is strictly larger than this bound. 
As such, CIs based on these methods are not tight. 


\subsection{Limitations of DRL}
We end this section by discussing the limitations of DRL. As we commented in Section \ref{secDROPE}, the validity of the Wald-type CI based on DRL requires both nuisance function estimators to converge at a rate faster than $(nT)^{-1/4}$. When this assumption is violated, the resulting CI cannot achieve nominal coverage. 

\begin{figure}[!t]
	\centering
	\includegraphics[width=0.55\textwidth]{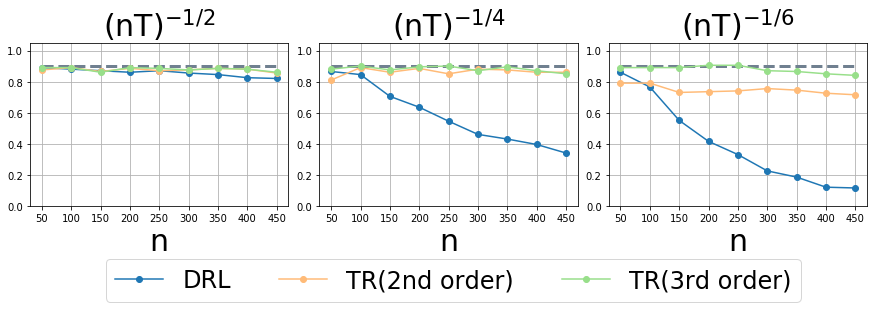}
	\caption{Empirical coverage probabilities for CIs based on DRL and the proposed triply-robust (TR) estimator, aggregated over 
		200 replications in the toy example. The nominal level is 90\% and $\gamma=0.95$. From left to right, we inject noises to the true Q-function and marginalized density ratio with standard errors proportional to $(nT)^{-1/2}$, $(nT)^{-1/4}$, and $(nT)^{-1/6}$, respectively. We vary the number of trajectories $n$ and fix $T=50$.  
	}
	\label{figure:toy_CI}
\end{figure}

To elaborate this, we design a toy example with three states (denote by A, B and C) arranged on a circle. The agent can move either clockwise or counter-clockwise. The reward is 1 if the agent reaches state A and 0 otherwise. We set the behaviour policy to a random policy. The target policy is very close to the optimal one. We inject some random errors to the true Q-function and marginalized density ratio to construct the CI based on DRL. It can be seen from Figure \ref{figure:toy_CI} that DRL is valid when the nuisance estimators are $(nT)^{-1/2}$-consistent but fails when they are $(nT)^{-1/4}$- or $(nT)^{-1/6}$-consistent. See Appendix \ref{sec:numerical_more_details} 
for details. 

We remark that the convergence rate assumption required by DRL is likely to be violated in complicated RL tasks with high-dimensional state variables. Take the Q-function estimator as an example. Suppose the true Q-function is H{\"o}lder smooth with exponent $\beta$ and $\widehat{Q}$ is computed via the deep Q-network \citep{mnih2015human} algorithm. Then similar to Theorem 4.4 of \citet{fan2020theoretical}, we can show that $\widehat{Q}$ converges at a rate of $(nT)^{-\beta/(2\beta+d)}$ where $d$ denotes the dimension of the state.   
When $d\ge 2\beta$, it is immediate to see that the assumption on $\widehat{Q}$ is violated. Learning the marginalized density ratio is even more challenging than learning the Q-function. We expect that the convergence rate assumption on $\widehat{\omega}$ would be violated as well. 

This motivates us to derive a valid CI under weaker and practically more feasible conditions. Our proposal requires to specify a hyper-parameter that 
determines the order of our value estimator. The larger this parameter, the weaker assumption our method requires. As an illustration, it can be seen from Figure \ref{figure:toy_CI} that our CI (denote by TR) achieves nominal coverage when the nuisance estimators are $(nT)^{-1/4}$- or even $(nT)^{-1/6}$-consistent. 

\section{Deeply-Debiased OPE}\label{secmethod}
\subsection{An Overview of  Our Proposal}\label{sec_triply_robust}
We first present an overview of our algorithm. Our procedure is composed of the following four steps, including data splitting, estimation of nuisance functions, debias iteration and construction of the CI.  

\textbf{Step 1. Data Splitting.}
We randomly divide the indices of all trajectories  $\{1,\cdots,n\}$ into $\mathbb{K}\ge 2$ disjoint subsets. Denote the $k$th subset by $\mathbb{I}_k$ and let  $\mathbb{I}_{k}^c=\{1,\cdots,n\}-\mathbb{I}_k$. 
Data splitting allows us to use one part of data ($\mathbb{I}_k^c$) to train RL models and the remaining part ($\mathbb{I}_k$) to do the estimation of the main parameter, i.e., $\eta^{\pi}$. We could aggregate the resulting estimates over different $k$ to get full efficiency. 
This allows us to establish the limiting distribution of the value estimator under minimal conditions. Data splitting has been commonly used in statistics and machine learning \citep[see e.g.,][]{chernozhukov2017double,kallus2019efficiently,shi2021testing}. 

\textbf{Step 2. Estimation of Nuisance Functions.}
This step is to estimate three nuisance functions, including the Q-function $Q^{\pi}$, the marginalized density ratio $\omega^{\pi}$, and a conditional density ratio $\tau^{\pi}$. Several algorithms in the literature can be applied to learn $Q^{\pi}$ and $\omega^{\pi}$, e.g., \citet{le2019batch,liu2018breaking,kallus2019efficiently,uehara2019minimax}. The conditional density ratio can be learned from the observed data in a similar fashion as $\omega^{\pi}$. See Section \ref{sec_nuisance} for more details. We use $\widehat{Q}_k$, $\widehat{\omega}_k$ and $\widehat{\tau}_k$ to denote the corresponding estimators, computed based on each data subset in $\mathbb{I}_k^c$. 

\textbf{Step 3. Debias Iteration.} 
This step is the key to our proposal. It recursively reduces the biases of the initial Q-estimator, allowing us to derive a valid CI for the target value under weaker and more practically feasible conditions on the estimated nuisance functions. Specifically, our CI allows the nuisance function estimator to converge at arbitrary rates. See Section \ref{secdebias} for details. 

\textbf{Step 4. Construction of the CI.}
Based on the debiased Q-estimator obtained in Step 3, we construct our value estimate and obtain a consistent estimator for its variance. A Wald-type CI can thus be derived. See Section \ref{secconCI} for details. 

In the following, we detail some major steps. We first introduce the debias iteration, as it contains the main idea of our proposal. We next detail Step 2 and Step 4. 

\subsection{Debias Iteration}\label{secdebias}

\subsubsection{The intuition for debias}
To motivate the proposed debias iteration, let us take a deeper look at DRL. 
Note that the second term on the right-hand-side of \eqref{term} is a plug-in estimator of the value based on the initial Q-estimator. The first term corresponds to an augmentation term. The purpose of adding this term is to offer additional protection against potential model misspecification of the Q-function. The resulting estimator's consistency relies on either $Q^{\pi}$ or $\omega^{\pi}$ to be correctly specified. As such, \eqref{term} can be understood as a de-biased version of the plug-in value estimator $\Mean_{s \sim \mathbb{G}, a \sim \pi(\cdot|s)}\widehat{Q}(a, s)$. 

Similarly, we can debias the initial Q-estimator $\widehat{Q}(a_0,s_0)$ for any $(a_0,s_0)$. Toward that end, we introduce the conditional density ratio. 
Specifically, by setting $\mathbb{G}(\bullet)$ to a Dirac measure $\mathbb{I}(\bullet=s_0)$ and further conditioning on an initial action $a_0$, the marginalized density ratio in \eqref{eqn:omega} becomes a conditional density ratio $\tau^{\pi}(a,s,a_0,s_0)$, defined as 
\begin{align*}
\frac{(1-\gamma)\{\mathcal{I}(a=a_0,s=s_0)+\sum_{t=1}^{+\infty}\gamma^t p_t^{\pi}(a,s|a_0,s_0)\} }{p_{\infty}(a,s)},
\end{align*}
where $p_t^{\pi}(a,s|a_0,s_0)$ denotes the probability of $(A_t,S_t)=(a,s)$ following policy $\pi$ conditional on the event that $\{A_0=a_0, S_0=s_0\}$, 
and $\mathcal{I}(\cdot)$ denotes the indicator function. By definition, the numerator corresponds to the discounted conditional visitation probability following $\pi$ given that the initial state-action pair equals $(s_0,a_0)$. In addition, one can show that $\omega^{\pi}(a,s)=\Mean_{s_0\sim \mathbb{G},a_0\sim \pi(\cdot|s_0)} \tau^{\pi}(a,s,a_0,s_0)$.  

By replacing $\widehat{\omega}_k$ in \eqref{term} with some estimated conditional density ratio $\widehat{\tau}_k$, we obtain the following estimation function
\begin{eqnarray}\label{eqn:debiasterm}
\begin{split}
\mathcal{D}_k^{(i,t)}Q(a,s)=Q(a,s)+ 
\frac{1}{1-\gamma}\widehat{\tau}_{k}(A_{i,t},S_{i,t},a,s)
\{R_{i,t} +\gamma
\Mean_{a' \sim \pi(\cdot|S_{i,t+1})}
Q(a',S_{i,t+1})- Q(A_{i,t},S_{i,t})\},  
\end{split}	
\end{eqnarray}
for any $Q$. Here, we refer $\mathcal{D}_k^{(i,t)}$ as the \textit{individual debiasing operator}, since it debiases any $Q$ based on an individual data tuple $(S_{i,t},A_{i,t},R_{i,t},S_{i,t+1})$. 

Similar to \eqref{term}, the augmentation term in \eqref{eqn:debiasterm} is to offer protection against potential model misspecification of the Q-function. As such, $\mathcal{D}_k^{(i,t)}Q(a,s)$ is unbiased to $Q^{\pi}(a,s)$ whenever $Q=Q^{\pi}$ or $\widehat{\tau}_k=\tau^{\pi}$. 

%

\subsubsection{The two-step debias iteration}
Based on the above discussions, 
a debiased version of the Q-estimator is given by averaging $\mathcal{D}_k^{(i,t)} \widehat{Q}_k$ over the data tuples in $\mathbb{I}_k$, i.e.,  $$\widehat{Q}^{(2)}_k=\frac{1}{|\mathbb{I}_k|T}\sum_{i\in \mathbb{I}_k}\sum_{0\le t<T}\mathcal{D}_k^{(i,t)} \widehat{Q}_k.$$ The bias of this estimator will decay at a faster rate than the initial Q-estimator $\widehat{Q}_k$, as shown in the following lemma. 
\begin{lemma}\label{lemma3}
	For any $k$, suppose $\widehat{Q}_k$ and $\widehat{\tau}_k$ converge in $L_2$-norm to $Q^{\pi}$ and $\tau^{\pi}$ at a rate of $(nT)^{-\alpha_1}$ and $(nT)^{-\alpha_2}$, respectively. 
	With weakly dependent data (see Condition (A1) in Section \ref{secTheory} in detail), we have
	\begin{eqnarray*}
		\Mean_{(a,s)\sim p_{\infty}} |\Mean \widehat{Q}_k^{(2)}(a,s)-Q(a,s)|=O\{(nT)^{-(\alpha_1+\alpha_2)}\}.
	\end{eqnarray*} 
\end{lemma}
To save space, we defer the detailed definition of $L_2$-norm convergence rate in Appendix \ref{sec:l2conv}. Suppose the square bias and variance of $\widehat{Q}_k$ are of the same order. Then we can show that the aggregated bias $\Mean_{(a,s)\sim p_{\infty}} |\Mean \widehat{Q}_k(a,s)-Q(a,s)|$ decays at a rate of $(nT)^{-\alpha_1}$. Consequently, Lemma \ref{lemma3} implies that the bias of $\widehat{Q}_k^{(2)}$ decays faster than $\widehat{Q}_k$. 

\begin{figure}[!t]
	\centering
	\includegraphics[width=0.48\textwidth]{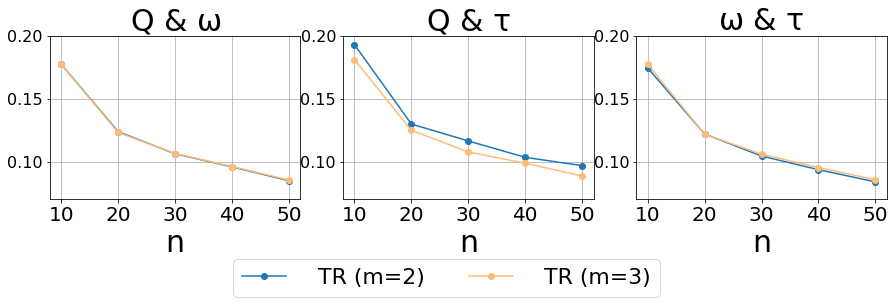}
	\caption{Root mean squared error (RMSE) of the proposed estimators in the toy example, computed over 200 replications. From left to right, we inject non-degenerate noises into $Q^{\pi}$ and $\omega^{\pi}$, $Q^{\pi}$ and $\tau^{\pi}$, $\omega^{\pi}$ and $\tau^{\pi}$, respectively. It can be seen that the RMSE decays as the sample size increases, when one of the three models is correctly specified. 
	}
	\label{figure:toy_triply}
\end{figure}

To illustrate the usefulness of the debiased Q-estimator $\widehat{Q}_k^{(2)}$, we propose to construct an estimating function $\psi_{i,t}^{(2)}$ for any $(i,t)\in \mathbb{I}_k$ by replacing $\widehat{Q}$ in \eqref{term} with $\widehat{Q}^{(2)}_k$. 
This yields our second-order estimator 
$$\widehat{\eta}^{(2)}_{\tiny{\textrm{TR}}}=(nT)^{-1} \sum_{i,t} \psi_{i,t}^{(2)}.$$ 

As we will show in Theorem \ref{thm_TR}, the proposed estimator $\widehat{\eta}^{(2)}_{\tiny{\textrm{TR}}}$ converges to the true value when one model for $Q^{\pi}$, $\omega^{\pi}$ or $\tau^{\pi}$ is correctly specified. As such, it is \textit{triply-robust}. See Figure \ref{figure:toy_triply} as an illustration. In addition, similar to Lemma \ref{lemma3}, the bias of $\widehat{\eta}^{(2)}_{\tiny{\textrm{TR}}}$ decays at a  faster rate than the DRL estimator. Specifically, we have the following results. 
\begin{lemma}\label{lemma4}
	Suppose the conditions in Lemma \ref{lemma3} hold and $\widehat{\omega}_k$ converges in $L_2$-norm to $\omega^k$ at a rate of $(NT)^{-\alpha_3}$ for any $k$. Let $\alpha=\min(1,\alpha_1+\alpha_2+\alpha_3)$. Then
	\begin{eqnarray*}
		|\Mean \widehat{\eta}^{(2)}_{\tiny{\textrm{TR}}}-\eta^{\pi}|=O\{(nT)^{-\alpha}\}.
	\end{eqnarray*}
\end{lemma}
In contrast, the bias of the DRL estimator decays at a rate of $(nT)^{-\alpha_1-\alpha_3}$. To ensure the resulting CI achieves valid coverage, we require the bias to decay at a rate faster than its variance which is typically of the order $O\{(nT)^{-1/2}\}$. As such, DRL requires $\min(\alpha_1,\alpha_3)>1/4$ whereas our second-order triply robust estimator relaxes this condition by requiring $\min(\alpha_1,\alpha_2,\alpha_3)>1/6$, as shown in Figure \ref{figure:toy_CI}. 

\subsubsection{The $m$-step debias iteration}
To further relax the convergence rate requirement, we can iteratively debias the Q-estimator to construct higher-order value estimates. Specifically, for any order $m\ge 2$, we iteratively apply the debiasing operator to the initial Q-estimator $m-1$ times and average over all individual tuples, leading to the following estimator,  
\begin{eqnarray*}
	\widehat{Q}^{(m)}_k={{|\mathbb{I}_k|T \choose (m-1)}}^{-1}\sum \mathcal{D}_k^{(i_1,t_1)}\cdots \mathcal{D}_k^{(i_{m-1},t_{m-1})} \widehat{Q}_k.
\end{eqnarray*}
Here, the sum is taken over all possible combinations of disjoint tuples $(i_1,t_1),(i_2,t_2),\cdots,(i_{m-1},t_{m-1})$ in the set $\{(i,t):i\in \mathbb{I}_k,0\le t<T\}$. Note that the definition involves repeated compositions of debiasing operator. For $m=3$, we present the detailed form in the appendix. In general, $\widehat{Q}_k^{(m)}(a,s)$ corresponds to an order $(m-1)$ U-statistic \citep[see e.g.,][]{lee2019u} for any $(a,s)$. 
The resulting value estimator $\widehat{\eta}^{(m)}_{\tiny{\textrm{TR}}}$ is given by $(nT)^{-1}\sum_{i,t} \psi_{i,t}^{(m)}$ where for any $(i,t)\in \mathbb{I}_k$, the estimating function $\psi_{i,t}^{(m)}$ is obtained by replacing $\widehat{Q}$ in \eqref{term} with $\widehat{Q}_k^{(m)}$. 

We make a few remarks. First, when $m=1$, $\widehat{Q}_k^{(m)}$ corresponds to the initial Q-estimator. As such, the proposed estimator reduces to the DRL estimator. When $m=2$, the definition here is consistent to the second-order triply-robust estimator. 

Second, 
for large $m$, calculating $\widehat{Q}_k^{(m)}$ is computationally intensive. In practice, we may approximate it using the incomplete U-statistics \cite{lee2019u,chen2019randomized} to facilitate the computation. 
For instance, to calculate $\widehat{Q}^{(3)}_{-k}(a,s)$, we could approximate it by 
averaging 
$\widehat{\mathcal{D}}^{(i_1,t_1)}_{k} \widehat{\mathcal{D}}^{(i_2,t_2)}_{k} \widehat{Q}_{k}(a,s)$
over $M$ pairs sampled from the set $\{(i_1,t_1,i_2,t_2):i_1,i_2\in \mathbb{I}_k, (i_1,t_1)\neq (i_2,t_2) \}$. 
We require $M$ to diverge with $nT$ 
such that the approximation error is asymptotically negligible. 
The computational complexity of our whole algorithm is analyzed in Appendix \ref{sec_computational_complexity} in the supplement. 

Third, the bias of the Q-estimator and that of the resulting value decrease as the order $m$ increases. Specifically, we have the following results. 
\begin{lemma}\label{lemma5}
	Suppose the conditions in Lemma \ref{lemma4} hold. Let $\alpha_*=\alpha_1+(m-1)\alpha_2$ and  $\alpha=\min(1,\alpha_1+(m-1)\alpha_2+\alpha_3)$. Then $\Mean_{(a,s)\sim p_{\infty}} |\Mean \widehat{Q}_k^{(m)}(a,s)-Q(a,s)|=O\{(nT)^{-(\alpha_*)}\}$ and $|\Mean \widehat{\eta}^{(m)}_{\tiny{\textrm{TR}}}-\eta^{\pi}|=O\{(NT)^{-\alpha}\}$. 
\end{lemma}
To ensure $\alpha<1/2$, it suffices to require $\alpha_1+(m-1)\alpha_2+\alpha_3>1/2$. As long as $\alpha_1,\alpha_2,\alpha_3>0$, there exists some $m$ that satisfies this condition. As such, the resulting bias decays faster than $(nT)^{-1/2}$. This yields the flexibility of our estimator as it allows the nuisance function estimator to converge at an arbitrary rate. When $m=2$,  Lemmas \ref{lemma3} and \ref{lemma4} are directly implied by Lemma \ref{lemma5}. 

\subsection{Learning the Nuisance Functions}
\label{sec_nuisance}
This step is to estimate the nuisance functions used in our algorithm, including $Q^{\pi}$, $\omega^{\pi}$, and  $\tau^{\pi}$, based on each data subset $\mathbb{I}_k^c$, for $k=1,\cdots,\mathbb{K}$.  

\textbf{The Q-function.} There are multiple learning methods available to produce an initial estimator for $Q^{\pi}$. We employ the fitted Q-evaluation method \citep{le2019batch} in our implementation. Based on the Bellman equation for $Q^{\pi}$ \citep[see Equation (4.6),][]{sutton2018reinforcement}, it iteratively solves the following optimization problem,
\begin{eqnarray*}
	\begin{split}
		\widehat{Q}^{{\ell}}=\argmin_{Q} 
		\sum_{\substack{i\in \mathbb{I}_k^c}}\sum_{t<T}
		\{
		\gamma \Mean_{a' \sim \pi(\cdot| S_{i, t+1})} \widehat{Q}^{\ell-1}(a',S_{i, t+1})
		+R_{i,t}- Q(A_{i, t}, S_{i, t})  \}^2,
	\end{split}	
\end{eqnarray*}
for $\ell=1,2,\cdots$, until convergence. 

We remark that the above optimization problem can be conveniently solved via supervised learning algorithms. In our experiments, we use random forest  \citep{breiman2001random} to estimate $Q^{\pi}$. 

\textbf{The Marginalized Density Ratio.} We next discuss the method for learning $\omega^{\pi}$. In our implementation, we employ the method of \citet{uehara2019minimax}. 
The following observation forms the basis of the method: when the process $\{(S_t,A_t)\}_{t\ge 0}$ is stationary, $\omega^{\pi}$ satisfies the equation $\Mean L(\omega^{\pi},f)=0$ for any function $f$, where $L(\omega^{\pi},f)$ equals 
\begin{eqnarray}\label{eqn_omega}
\begin{split}
\Big[\Mean_{a \sim \pi(\cdot|S_{t+1})} \{\omega^{\pi}(A_{t},S_{t})
(\gamma f(a, S_{t+1})- f(A_{t},S_{t}) ) \}
+ (1-\gamma) \Mean_{s \sim \mathbb{G}, a \sim \pi(\cdot|s)} f(a, s). 
\end{split}
\end{eqnarray} 
As such, $\omega^{\pi}$ can be learned by solving the following mini-max problem, 
\begin{eqnarray}\label{eqn:solveL}
\argmin_{\omega\in \Omega} \sup_{f\in \mathcal{F}} \{\Mean L(\omega, f)\}^2, 
\end{eqnarray}
for some function classes $\Omega$ and $\mathcal{F}$. 
The expectation in \eqref{eqn:solveL} is approximated by the sample mean. To simplify the calculation, we choose $\mathcal{F}$ to be a reproducing kernel Hilbert space (RKHS). This yields a closed form expression for $\sup_{f\in \mathcal{F}} \{\Mean L(\omega,f)\}^2$. Consequently,  $\omega^{\pi}$ can be learned by solving the outer minimization via stochastic gradient descent. 
To save space, we defer the details to Appendix \ref{secomega} in the supplementary article. 

\textbf{The Conditional Density Ratio.} Finally, we 
develop a method to learn $\tau^{\pi}$ based on the observed data. 
Note that $\tau^{\pi}$ can be viewed as a version of $\omega^{\pi}$ by conditioning on the initial state-action pair. Similar to \eqref{eqn_omega}, we have 
\begin{eqnarray*}
	\begin{split}
		\Mean\Big[\Mean_{a \sim \pi(\cdot|S_{t+1})} \tau^{\pi}(A_{t},S_{t},a_0,s_0)
		\{\gamma g(a, S_{t+1})
		- g(A_{t},S_{t}) \}\Big]
		+ (1-\gamma) g(a_0, s_0)=0, 
	\end{split}
\end{eqnarray*}
for any $g$ and state-action pair $(a_0,s_0)$, or equivalently, 
\begin{eqnarray}\label{eqn_conomega}
\begin{split}
\Mean\Big[\Mean_{a \sim \pi(\cdot|S_{t+1})} \tau^{\pi}(A_{t},S_{t},a_0,s_0)
\{\gamma f(a, S_{t+1},a_0,s_0)
- f(A_{t},S_{t},a_0,s_0 ) \}\Big]
+ (1-\gamma) f(a_0, s_0,a_0,s_0)=0, 
\end{split}
\end{eqnarray}
for any function $f$ and $(a_0,s_0)$. Integrating $(a_0,s_0)$ on the left-hand-side of \eqref{eqn_conomega} with respect to the stationary state-action distribution $p_{\infty}$, we obtain the following lemma.  
\begin{lemma}\label{lemma_omegastar}
	Suppose the process $\{(A_t,S_t)\}_{t\ge 0}$ is strictly stationary. 
	For any function $f$, $\tau^{\pi}$ satisfies the equation $h(\tau^{\pi},f)=0$ where $h(\tau^{\pi},f)$ is given by
	\begin{eqnarray*}
		\begin{split}	
			&\Mean \Big[(1-\gamma)  f(A_{i_1,t_1},S_{i_1,t_1},A_{i_1,t_1},S_{i_1,t_1})-
			\tau^{\pi}(A_{i_2,t_2},S_{i_2,t_2},A_{i_1,t_1},S_{i_1,t_1})\times \\&\{f(A_{i_2,t_2},S_{i_2,t_2},A_{i_1,t_1},S_{i_1,t_1})
			-\gamma 	\Mean_{a \sim \pi(\cdot|S_{i_2,t_2+1})} f(S_{i_2,t_2+1},a;S_{i_1,t_1},A_{i_1,t_1})\}\Big],
		\end{split}	
	\end{eqnarray*}
	for any $i_1\neq i_2$ such that  $(S_{i_1,t_1},A_{i_1,t_1},S_{i_1,t_1+1})$ and $(S_{i_2,t_2},A_{i_2,t_2},S_{i_2,t_2+1})$ are independent. 
\end{lemma}
Similar to Lemma 15 of \citet{kallus2019efficiently}, we can also show that $\tau^\pi$ is the only function that satisfies Lemma \ref{lemma_omegastar}.
Motivated by this lemma, $\tau^\pi$ can be learned by solving the following mini-max optimization problem
\begin{eqnarray}\label{optimize}
\argmin_{\tau\in \mathcal{T}} \sup_{f\in \mathcal{F}} h^2(\tau,f), 
\end{eqnarray}
for some function classes $\mathcal{T}$ and $\mathcal{F}$. For any $\tau$ and $f$, we estimate $h(\tau,f)$ based on the observed data. Setting $\mathcal{F}$ to an RKHS and $\mathcal{T}$ to a class of deep neural networks, the above optimization can be solved in a similar fashion as \eqref{eqn:solveL}. We defer the details to Appendix \ref{secomegastar} to save space.

\subsection{Construction of the CI}\label{secconCI}
In this step, we construct a CI based on $\widehat{\eta}_{\textrm{TR}}^{(m)}$. Specifically, under mild assumptions, the asymptotic variance of $\sqrt{nT}\widehat{\eta}_{\textrm{TR}}^{(m)}$ can be consistently estimated by the sampling variance estimator of $\{\psi_{i,t}^{(m)}\}_{i,t}$ (denote by  $\{\widehat{\sigma}^{(m)}\}^2$). 
For a given significance level  $\alpha$, the corresponding two-sided CI is given by $[\widehat{\eta}_{\textrm{TR}}^{(m)} - z_{\alpha/2} (nT)^{-1/2}	\widehat{\sigma}^{(m)} ,\widehat{\eta}_{\textrm{TR}}^{(m)}+z_{\alpha/2} (nT)^{-1/2}	\widehat{\sigma}^{(m)}]$ where $z_{\alpha}$ corresponds to the upper $\alpha$th quantile of a standard normal random variable. 

\section{Robustness, Efficiency and Flexibility}
\label{secTheory}

We first summarize our results. Theorem \ref{thm_TR} establishes the triply-robust property of our value estimator $\widehat{\eta}^{(m)}$. Theorem \ref{thm_eff} shows the asymptotic variance of $\widehat{\eta}^{(m)}$ achieves the semiparametric efficiency bound \eqref{lower_bound}. As such, our estimator is sample efficient. Theorem \ref{thm_adaptive} implies that our CI achieves nominal coverage under 
weaker and much practically feasible conditions than DRL. All of our theoretical guarantees are derived under the asymptotic framework that requires either the number of trajectories $n$ or the number of decision points $T$ per trajectory to diverge to infinity. Results of this type provide useful theoretical guarantees for different types of applications, and are referred as bidirectional theories.

We next introduce some conditions.

(A1) The process $\{(S_t,A_t,R_t)\}_{t\ge 0}$ is strictly stationary and exponentially $\beta$-mixing \citep[see e.g.,][for a detailed explanation of this definition]{bradley2005basic}. 

(A2) For any $k$, $\widehat{Q}_k$, $\widehat{\tau}_k$ and $\widehat{\omega}_k$ converge in $L_2$-norm to $Q^{\pi}$, $\tau^{\pi}$ and $\omega^{\pi}$ at a rate of $(nT)^{-\alpha_1}$, $(nT)^{-\alpha_2}$ and $(nT)^{-\alpha_3}$ for any $\alpha_1,\alpha_2$ and $\alpha_3>0$, respectively. 

(A3) $\tau^{\pi}$ and $\omega^{\pi}$ are uniformly bounded away from infinity. 

Condition (A1) allows the data observations to be weakly dependent. %
When the behavior policy is not history-dependent, the process $\{(S_t,A_t,R_t)\}_{t\ge 0}$ forms a Markov chain. 
The exponential $\beta$-mixing condition is automatically satisfied when the Markov chain is geometrically ergodic \citep[see Theorem 3.7 of][]{bradley2005basic}. Geometric ergodicity is less restrictive than those  imposed in the existing reinforcement learning literature that requires observations to be  independent \citep[see e.g.,][]{dai2020coindice} or to follow a uniform-ergodic Markov chain \citep[see e.g.,][]{bhandari2018finite,zou2019}. We also remark that the stationarity assumption in (A1) is assumed for convenience, since the Markov chain will eventually reach stationarity. 

Condition (A2) characterizes the theoretical requirements on the nuisance function estimators. This assumption is mild as we require these estimators to converge at any rate. When using kernels or neural networks for function approximation, the corresponding convergence rates of $\widehat{Q}_k$ and $\widehat{\omega}_k$ are provided in \citet{fan2020theoretical,liao2020batch}. The convergence rate for $\widehat{\tau}_k$ can be similarly derived as $\widehat{\omega}_k$. 

Condition (A3) essentially requires that any state-action pair supported by the density function $(1-\gamma)\sum_{t\ge 0} \gamma^t p_t^{\pi}$ is supported by the stationary behavior density function as well. This assumption is similar to the sequential overlap condition imposed by \citet{kallus2020double}.

\begin{thm}[Robustness]\label{thm_TR}
	Suppose (A1) and (A3) hold, and $\widehat{Q}_k$, $\widehat{\tau}_k$, $\widehat{\omega}_k$ are uniformly bounded away from infinity almost surely. 
	Then for any $m$, as either $n$ or $T$ diverges to infinity, our value estimator $\widehat{\eta}^{(m)}_{\tiny{\textrm{TR}}}$ is consistent when $\widehat{Q}_k$, $\widehat{\tau}_k$ or $\widehat{\omega}_k$ converges in $L_2$-norm to $Q^{\pi}$, $\tau^{\pi}$ or $\omega^{\pi}$ for any $k$. 
\end{thm}
Theorem 1 does not rely on Condition (A2). It only requires one of the three nuisance estimators to converge. As such, it is more robust than existing doubly-robust estimators. 

\begin{thm}[Efficiency]\label{thm_eff}
	Suppose (A1) and (A2) hold, and $\widehat{Q}_k$, $\widehat{\tau}_k$, $\widehat{\omega}_k$, $\tau^{\pi}$, $\omega^{\pi}$ are uniformly bounded away from infinity almost surely. Then for any $m$, as either $n$ or $T$ approaches infinity, $\sqrt{nT}(\widehat{\eta}_{\tiny{\textrm{TR}}}^{(m)}-\Mean \widehat{\eta}_{\tiny{\textrm{TR}}}^{(m)})\stackrel{d}{\to} N(0,\sigma^2)$ where $\sigma^2$ corresponds to the efficiency bound in \eqref{lower_bound}.
\end{thm}

We make some remarks. In the proof of Theorem \ref{thm_eff}, we show that $\widehat{\eta}_{\tiny{\textrm{TR}}}^{(m)}$ is asymptotically equivalent to an $m$th order U-statistic. According to the Hoeffding decomposition \citep{hoeffding1948class}, we can decompose the U-statistic into the sum $\eta^{\pi}+\sum_{j=1}^m \widehat{\eta}_j$, 
where $\eta^{\pi}$ is the main effect term that corresponds to the asymptotic mean of the value estimator, $\widehat{\eta}_1$ is the first-order term 
\begin{eqnarray*}
	\frac{1}{nT(1-\gamma)}\sum_{i=1}^n\sum_{t=0}^{T-1}\omega^{\pi}(A_{i,t},S_{i,t})\{R_{i,t}+\gamma \Mean_{a\sim \pi(\cdot|S_{i,t+1})}Q^{\pi}(a,S_{i,t+1})-Q^{\pi}(A_{i,t},S_{i,t})\},
\end{eqnarray*}
and $\widehat{\eta}_j$ corresponds to a $j$th order degenerate U-statistic for any $j\ge 2$. See Part 3 of the proof of Theorem \ref{thm_eff} for details. 
Note that the DRL estimator is asymptotically equivalent to $\eta^{\pi}+\widehat{\eta}_1$.  Under (A1), these $\widehat{\eta}_j$s are asymptotically uncorrelated. As such, the variance of our estimator is asymptotically equivalent to
\begin{eqnarray*}
	\sum_{j=1}^m \Var(\widehat{\eta}_j)=\sum_{j=1}^m {nT\choose j}^{-1}\sigma_j^2,
\end{eqnarray*}
where $\sigma_j^2$s are bounded. When $j=1$, we have $\sigma_j^2=\sigma^2$. For $j\ge 2$, $\Var(\widehat{\eta}_j)$ decays at a faster rate than $\Var(\widehat{\eta}_1)=\sigma^2(nT)^{-1}$. As such, the variance of our estimator is asymptotically equivalent to that of DRL. 

However, in finite sample, the variance of the proposed estimator is strictly larger than DRL, due to the presence of high-order variance terms. This is consistent with our experiment results (see Section \ref{secExperiments}) where we find the proposed CI is usually slightly wider than that based on DRL. This reflects a bias-variance trade-off. Specifically, our procedure alleviates the bias of the DRL estimator to obtain valid uncertainty quantification. The resulting estimator would have a strictly larger variance than DRL in finite samples, although the difference is asymptotically negligible. We also remark that in interval estimation, the first priority is to ensure the CI has nominal coverage. This requires an estimator’s bias to decay faster than its variance. The second priority is to shorten the length of CI (the variance of the estimator) if possible. In that sense, variance is less significant than bias.


\begin{thm}[Flexibility]\label{thm_adaptive}
	Suppose the conditions in Theorem \ref{thm_eff} hold. Then as long as $m$ satisfies $\alpha_1+(m-1)\alpha_2+\alpha_3>1/2$, the proposed CI achieves nominal coverage. 
\end{thm}

Theorem \ref{thm_adaptive} implies that our CI allows the nuisance functions to diverge at an arbitrary rate for sufficiently large $m$. 



\section{Experiments}\label{secExperiments}
\begin{figure*}[!t]
	\centering
	\includegraphics[width=0.79\textwidth]{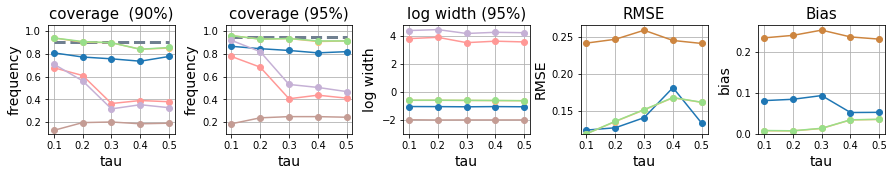} \\
	\includegraphics[width=0.79\textwidth]{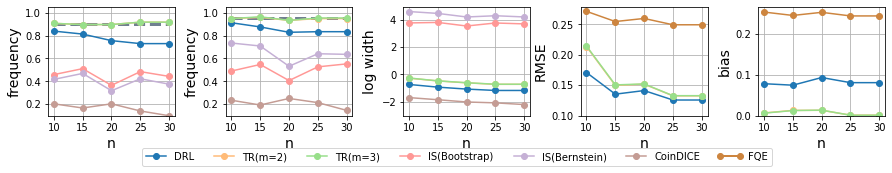} 
	\caption{Results for Cartpole. We fix $n = 20$ and vary $\tau$ in the upper subplots, and fix $\tau = 0.3$ and vary $n$ in the lower subplots. The subplots from left to right are about the coverage frequency with $\alpha = 0.9$, the coverage frequency with $\alpha = 0.95$, the mean log width of CIs with $\alpha = 0.95$, the RMSE of value estimates, and the bias of value estimates, respectively. The yellow line (TR, $m=2$) and green line (TR, $m=3$) are largely overlapped.}
	\label{figure:Cartpole}
\end{figure*}
\begin{figure*}[!h]
	\centering
	\includegraphics[width=0.79\textwidth]{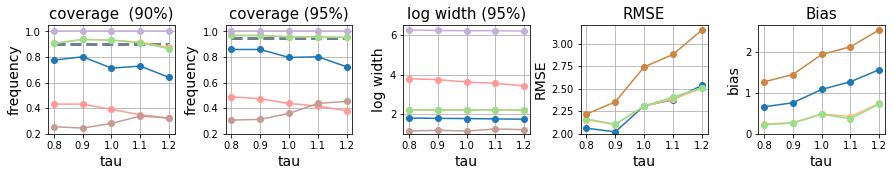} \\
	\includegraphics[width=0.79\textwidth]{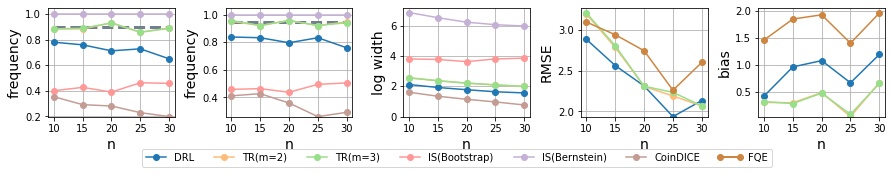} 
	\caption{Results for Diabetes. We fix $n = 20$ and vary $\tau$ in the upper subplots, and fix $\tau = 1.0$ and vary $n$ in the lower subplots. 
		Same legend as Figure \ref{figure:Cartpole}. The yellow line (TR, $m=2$) and green line (TR, $m=3$) are largely overlapped.}
	\label{figure:Diabetes}
\end{figure*}

In this section, we evaluate the empirical performance of our method using two synthetic datasets: CartPole from the OpenAI Gym environment \cite{brockman2016openai} 
and a simulation environment (referred to as Diabetes) to simulate the OhioT1DM data \citep{shi2020does}. In the second environment, the goal is to learn an optimal policy as a function of patients' time-varying covariates to improve their health status. 
In both settings, 
following \citet{uehara2019minimax}, 
we first learn a near-optimal policy as the target policy, 
and then apply softmax on its Q-function divided by a temperature parameter $\tau$ to set the action probabilities to define a behaviour policy.  
A larger $\tau$ implies a larger difference between the behaviour policy and the target policy.

We denote the proposed method as TR and present results with $m = 2$ and $3$. The choice of $m$ represents a trade-off. In theory, $m$ shall be as large as
possible to guarantee the validity of our CI. Yet, the computation complexity increases exponentially in $m$. 
In our experiments, we find that setting $m=3$ yields satisfactory performance in general. 

For point estimation, we compare the bias and RMSE of our method with DRL and the estimator computed via  fitted-Q evaluation (FQE). 
For interval estimation, we compare the proposed CI with several competing baselines, including 
CoinDICE \citep{dai2020coindice},  stepwise IS-based estimator with bootstrapping \cite{thomas2015high}, stepwise IS-based estimator with Bernstein inequality \cite{thomas2015high2}, and the CI based on DRL. 
For each method, we report the empirical coverage probability and the average length of the constructed CI.

We set $T = 300$ and $\gamma = 0.98$ for CartPole, and $T = 200$ and $\gamma = 0.95$ for Diabetes. 
For both environments, we vary the number of trajectories $n$ and the temperature $\tau$ to design different settings. Results are aggregated over 200 replications. 
Note that FQE and DR share the same subroutines with TR, and hence the same hyper-parameters are used. 
More details about the environments and the implementations  
can be found in Section \ref{sec:numerical_more_details} of the supplement. 

The results for CartPole and Diabetes are depicted in 
Figures \ref{figure:Cartpole} and \ref{figure:Diabetes}, respectively. 
We summarize our findings as follows. 
In terms of interval estimation, first, the proposed CI achieves nominal coverage in all cases, whereas the CI based on DRL fails to cover the true value. This demonstrates that the proposed method is more robust than DRL. In addition, the average length of our CI is slightly larger than that of DRL in all cases. This reflects the bias-variance tradeoff we detailed in Section \ref{secTheory}. 
Second, CoinDice 
yields the narrowest CI. However, its empirical coverage probability is well below the nominal level in all cases. As we have commented in the introduction, this is due to that their method requires i.i.d. observations and would fail with weakly dependent data. Please refer to Appendix \ref{sec:coindice} for details. Third, the stepwise IS-based estimators suffer from the curse of horizon. 
The lengths of the resulting CIs are much larger than ours. Moreover, the CI based on bootstrapping the stepwise IS-estimator fails to achieve nominal coverage. This is because the standard bootstrap method is not valid with weakly dependent data. 

In terms of point estimation, TR yields smaller bias than DRL in all cases. FQE suffers from the largest bias among the three methods. 
The RMSEs of DRL and TR are comparable and generally smaller than that of FQE. This demonstrates the efficiency of the proposed estimator. 
\section{Discussion}
\subsection{Order Selection}
In this paper, we develop a deeply-debiased procedure for off-policy interval estimation. Our proposal relies on the specification of $m$, the number of the debias iteration. The choice of $m$ represents a trade-off. In theory, $m$ shall be as large as possible to reduce the bias of the value estimator and guarantee the validity of the resulting CI. Yet, the variance of the value estimator and the computation of our procedure increase with $m$. In the statistics literature, Lepski's method is a data-adaptive procedure for identifying optimal tuning parameter where cross-validation is difficult to implement, as in our setup \citep[see e.g.,][]{su2020adaptive}. It can be naturally coupled with the proposed method for order selection, to balance the bias-variance trade-off. Practical version of Lepski's method was developed using bootstrap in \citet{chernozhukov2014anti}. This idea is worthwhile to explore and we leave it for future research. 
\subsection{Nonasymptotic Confidence Bound}
Non-asymptotic confidence bound is typically obtained by applying concentration inequalities \citep[e.g., Hoeffding's inequality or Bernstein inequality][]{van1996weak} to a sum of uncorrelated variables. In our setup, the proposed estimator is a U-statistic. We could apply concentration inequalities to U-statistics \citep[see e.g.,][]{feng2020accountable} to derive the confidence bound. Alternatively, we may apply self-normalized moderate deviation inequalities \citep{pena2008self} to derive the non-asymptotic bound. The resulting confidence bound will be wider than the proposed CI. However, it is valid even with small sample size.
\subsection{Hardness of Learning of $\tau^{\pi}$}
Learning $\tau^{\pi}$ could be much challenging than $\omega^{\pi}$. In our current numerical experiments, all the state variables are continuous and it is challenging to obtain the ground truth of the conditional density ratio which involves estimation of a high-dimensional conditional density. As such, we did not investigate the goodness-of-fit of the proposed estimator for $\tau^{\pi}$. It would be practically interesting to explore the optimal neural network structure to approximate $\tau^{\pi}$ and investigate the finite-sample rate of convergence of our estimator. However, this is beyond the scope of the current paper. We leave it for future research. 
\subsection{Extension to Exploration}
Finally, we remark that based on the proposed debiased Q-estimator, a two-sided CI can be similarly to quantify its uncertainty. It allows us to follow the ``optimism in the face of uncertainty" principle for online exploration. This is another topic that warrants future investigation.  

\bibliography{D2OPE}

\begin{thebibliography}{60}
\providecommand{\natexlab}[1]{#1}
\providecommand{\url}[1]{\texttt{#1}}
\expandafter\ifx\csname urlstyle\endcsname\relax
  \providecommand{\doi}[1]{doi: #1}\else
  \providecommand{\doi}{doi: \begingroup \urlstyle{rm}\Url}\fi

\bibitem[Andrychowicz et~al.(2020)Andrychowicz, Baker, Chociej, Jozefowicz,
  McGrew, Pachocki, Petron, Plappert, Powell, Ray,
  et~al.]{andrychowicz2020learning}
Andrychowicz, O.~M., Baker, B., Chociej, M., Jozefowicz, R., McGrew, B.,
  Pachocki, J., Petron, A., Plappert, M., Powell, G., Ray, A., et~al.
\newblock Learning dexterous in-hand manipulation.
\newblock \emph{The International Journal of Robotics Research}, 39\penalty0
  (1):\penalty0 3--20, 2020.

\bibitem[Bhandari et~al.(2018)Bhandari, Russo, and Singal]{bhandari2018finite}
Bhandari, J., Russo, D., and Singal, R.
\newblock A finite time analysis of temporal difference learning with linear
  function approximation.
\newblock \emph{arXiv preprint arXiv:1806.02450}, 2018.

\bibitem[Bickel et~al.(1993)Bickel, Klaassen, Bickel, Ritov, Klaassen, Wellner,
  and Ritov]{bickel1993efficient}
Bickel, P.~J., Klaassen, C.~A., Bickel, P.~J., Ritov, Y., Klaassen, J.,
  Wellner, J.~A., and Ritov, Y.
\newblock \emph{Efficient and adaptive estimation for semiparametric models},
  volume~4.
\newblock Johns Hopkins University Press Baltimore, 1993.

\bibitem[Bradley(2005)]{bradley2005basic}
Bradley, R.~C.
\newblock Basic properties of strong mixing conditions. a survey and some open
  questions.
\newblock \emph{Probability Surveys}, 2:\penalty0 107--144, 2005.

\bibitem[Breiman(2001)]{breiman2001random}
Breiman, L.
\newblock Random forests.
\newblock \emph{Machine learning}, 45\penalty0 (1):\penalty0 5--32, 2001.

\bibitem[Brockman et~al.(2016)Brockman, Cheung, Pettersson, Schneider,
  Schulman, Tang, and Zaremba]{brockman2016openai}
Brockman, G., Cheung, V., Pettersson, L., Schneider, J., Schulman, J., Tang,
  J., and Zaremba, W.
\newblock Openai gym.
\newblock \emph{arXiv preprint arXiv:1606.01540}, 2016.

\bibitem[Chen et~al.(2019)Chen, Kato, et~al.]{chen2019randomized}
Chen, X., Kato, K., et~al.
\newblock Randomized incomplete $ u $-statistics in high dimensions.
\newblock \emph{Annals of Statistics}, 47\penalty0 (6):\penalty0 3127--3156,
  2019.

\bibitem[Chernozhukov et~al.(2014)Chernozhukov, Chetverikov, Kato,
  et~al.]{chernozhukov2014anti}
Chernozhukov, V., Chetverikov, D., Kato, K., et~al.
\newblock Anti-concentration and honest, adaptive confidence bands.
\newblock \emph{Annals of Statistics}, 42\penalty0 (5):\penalty0 1787--1818,
  2014.

\bibitem[Chernozhukov et~al.(2017)Chernozhukov, Chetverikov, Demirer, Duflo,
  Hansen, and Newey]{chernozhukov2017double}
Chernozhukov, V., Chetverikov, D., Demirer, M., Duflo, E., Hansen, C., and
  Newey, W.
\newblock Double/debiased/neyman machine learning of treatment effects.
\newblock \emph{American Economic Review}, 107\penalty0 (5):\penalty0 261--65,
  2017.

\bibitem[Dai et~al.(2020)Dai, Nachum, Chow, Li, Szepesvari, and
  Schuurmans]{dai2020coindice}
Dai, B., Nachum, O., Chow, Y., Li, L., Szepesvari, C., and Schuurmans, D.
\newblock Coindice: Off-policy confidence interval estimation.
\newblock \emph{Advances in neural information processing systems}, 33, 2020.

\bibitem[Dedecker \& Louhichi(2002)Dedecker and Louhichi]{dedecker2002maximal}
Dedecker, J. and Louhichi, S.
\newblock Maximal inequalities and empirical central limit theorems.
\newblock In \emph{Empirical process techniques for dependent data}, pp.\
  137--159. Springer, 2002.

\bibitem[Denker \& Keller(1983)Denker and Keller]{denker1983u}
Denker, M. and Keller, G.
\newblock On u-statistics and v. mise’statistics for weakly dependent
  processes.
\newblock \emph{Zeitschrift f{\"u}r Wahrscheinlichkeitstheorie und verwandte
  Gebiete}, 64\penalty0 (4):\penalty0 505--522, 1983.

\bibitem[Deshpande et~al.(2018)Deshpande, Mackey, Syrgkanis, and
  Taddy]{deshpande2018accurate}
Deshpande, Y., Mackey, L., Syrgkanis, V., and Taddy, M.
\newblock Accurate inference for adaptive linear models.
\newblock In \emph{International Conference on Machine Learning}, pp.\
  1194--1203. PMLR, 2018.

\bibitem[Duchi et~al.(2016)Duchi, Glynn, and Namkoong]{duchi2016statistics}
Duchi, J., Glynn, P., and Namkoong, H.
\newblock Statistics of robust optimization: A generalized empirical likelihood
  approach.
\newblock \emph{arXiv preprint arXiv:1610.03425}, 2016.

\bibitem[Fan et~al.(2020)Fan, Wang, Xie, and Yang]{fan2020theoretical}
Fan, J., Wang, Z., Xie, Y., and Yang, Z.
\newblock A theoretical analysis of deep q-learning.
\newblock In \emph{Learning for Dynamics and Control}, pp.\  486--489. PMLR,
  2020.

\bibitem[Farajtabar et~al.(2018)Farajtabar, Chow, and
  Ghavamzadeh]{farajtabar2018more}
Farajtabar, M., Chow, Y., and Ghavamzadeh, M.
\newblock More robust doubly robust off-policy evaluation.
\newblock \emph{arXiv preprint arXiv:1802.03493}, 2018.

\bibitem[Feng et~al.(2020)Feng, Ren, Tang, and Liu]{feng2020accountable}
Feng, Y., Ren, T., Tang, Z., and Liu, Q.
\newblock Accountable off-policy evaluation with kernel bellman statistics.
\newblock \emph{arXiv preprint arXiv:2008.06668}, 2020.

\bibitem[Hadad et~al.(2021)Hadad, Hirshberg, Zhan, Wager, and
  Athey]{hadad2021confidence}
Hadad, V., Hirshberg, D.~A., Zhan, R., Wager, S., and Athey, S.
\newblock Confidence intervals for policy evaluation in adaptive experiments.
\newblock \emph{Proceedings of the National Academy of Sciences}, 118\penalty0
  (15), 2021.

\bibitem[Hanna et~al.(2016)Hanna, Stone, and Niekum]{hanna2016bootstrapping}
Hanna, J.~P., Stone, P., and Niekum, S.
\newblock Bootstrapping with models: Confidence intervals for off-policy
  evaluation.
\newblock \emph{arXiv preprint arXiv:1606.06126}, 2016.

\bibitem[Hoeffding(1948)]{hoeffding1948class}
Hoeffding, W.
\newblock A class of statistics with asymptotically normal distribution.
\newblock \emph{The Annals of Mathematical Statistics}, pp.\  293--325, 1948.

\bibitem[Jiang \& Huang(2020)Jiang and Huang]{jiang2020minimax}
Jiang, N. and Huang, J.
\newblock Minimax value interval for off-policy evaluation and policy
  optimization.
\newblock \emph{Advances in Neural Information Processing Systems}, 33, 2020.

\bibitem[Jiang \& Li(2016)Jiang and Li]{jiang2016doubly}
Jiang, N. and Li, L.
\newblock Doubly robust off-policy value evaluation for reinforcement learning.
\newblock In \emph{International Conference on Machine Learning}, pp.\
  652--661. PMLR, 2016.

\bibitem[Kallus \& Uehara(2019)Kallus and Uehara]{kallus2019efficiently}
Kallus, N. and Uehara, M.
\newblock Efficiently breaking the curse of horizon in off-policy evaluation
  with double reinforcement learning.
\newblock \emph{arXiv preprint arXiv:1909.05850}, 2019.

\bibitem[Kallus \& Uehara(2020)Kallus and Uehara]{kallus2020double}
Kallus, N. and Uehara, M.
\newblock Double reinforcement learning for efficient off-policy evaluation in
  markov decision processes.
\newblock \emph{Journal of Machine Learning Research}, 21\penalty0
  (167):\penalty0 1--63, 2020.

\bibitem[Kitamura et~al.(1997)]{kitamura1997empirical}
Kitamura, Y. et~al.
\newblock Empirical likelihood methods with weakly dependent processes.
\newblock \emph{The Annals of Statistics}, 25\penalty0 (5):\penalty0
  2084--2102, 1997.

\bibitem[Le et~al.(2019)Le, Voloshin, and Yue]{le2019batch}
Le, H.~M., Voloshin, C., and Yue, Y.
\newblock Batch policy learning under constraints.
\newblock \emph{arXiv preprint arXiv:1903.08738}, 2019.

\bibitem[Lee(2019)]{lee2019u}
Lee, A.~J.
\newblock \emph{U-statistics: Theory and Practice}.
\newblock Routledge, 2019.

\bibitem[Liao et~al.(2020)Liao, Qi, and Murphy]{liao2020batch}
Liao, P., Qi, Z., and Murphy, S.
\newblock Batch policy learning in average reward markov decision processes.
\newblock \emph{arXiv preprint arXiv:2007.11771}, 2020.

\bibitem[Liu et~al.(2018)Liu, Li, Tang, and Zhou]{liu2018breaking}
Liu, Q., Li, L., Tang, Z., and Zhou, D.
\newblock Breaking the curse of horizon: Infinite-horizon off-policy
  estimation.
\newblock In \emph{Advances in Neural Information Processing Systems}, pp.\
  5356--5366, 2018.

\bibitem[Luedtke \& van~der Laan(2017)Luedtke and van~der
  Laan]{luedtke2017evaluating}
Luedtke, A.~R. and van~der Laan, M.~J.
\newblock Evaluating the impact of treating the optimal subgroup.
\newblock \emph{Statistical methods in medical research}, 26\penalty0
  (4):\penalty0 1630--1640, 2017.

\bibitem[Mackey et~al.(2018)Mackey, Syrgkanis, and Zadik]{mackey2018orthogonal}
Mackey, L., Syrgkanis, V., and Zadik, I.
\newblock Orthogonal machine learning: Power and limitations.
\newblock In \emph{International Conference on Machine Learning}, pp.\
  3375--3383. PMLR, 2018.

\bibitem[Marling \& Bunescu(2018)Marling and Bunescu]{marling2018ohiot1dm}
Marling, C. and Bunescu, R.~C.
\newblock The ohiot1dm dataset for blood glucose level prediction.
\newblock In \emph{KHD@ IJCAI}, pp.\  60--63, 2018.

\bibitem[Mnih et~al.(2015)Mnih, Kavukcuoglu, Silver, Rusu, Veness, Bellemare,
  Graves, Riedmiller, Fidjeland, Ostrovski, et~al.]{mnih2015human}
Mnih, V., Kavukcuoglu, K., Silver, D., Rusu, A.~A., Veness, J., Bellemare,
  M.~G., Graves, A., Riedmiller, M., Fidjeland, A.~K., Ostrovski, G., et~al.
\newblock Human-level control through deep reinforcement learning.
\newblock \emph{nature}, 518\penalty0 (7540):\penalty0 529--533, 2015.

\bibitem[Mukherjee et~al.(2017)Mukherjee, Newey, and
  Robins]{mukherjee2017semiparametric}
Mukherjee, R., Newey, W.~K., and Robins, J.~M.
\newblock Semiparametric efficient empirical higher order influence function
  estimators.
\newblock \emph{arXiv preprint arXiv:1705.07577}, 2017.

\bibitem[Murphy et~al.(2001)Murphy, van~der Laan, Robins, and
  Group]{murphy2001marginal}
Murphy, S.~A., van~der Laan, M.~J., Robins, J.~M., and Group, C. P. P.~R.
\newblock Marginal mean models for dynamic regimes.
\newblock \emph{Journal of the American Statistical Association}, 96\penalty0
  (456):\penalty0 1410--1423, 2001.

\bibitem[Nachum et~al.(2019)Nachum, Chow, Dai, and Li]{nachum2019dualdice}
Nachum, O., Chow, Y., Dai, B., and Li, L.
\newblock Dualdice: Behavior-agnostic estimation of discounted stationary
  distribution corrections.
\newblock In \emph{Advances in Neural Information Processing Systems}, pp.\
  2318--2328, 2019.

\bibitem[Owen(2001)]{owen2001empirical}
Owen, A.~B.
\newblock \emph{Empirical likelihood}.
\newblock CRC press, 2001.

\bibitem[Pe{\~n}a et~al.(2008)Pe{\~n}a, Lai, and Shao]{pena2008self}
Pe{\~n}a, V.~H., Lai, T.~L., and Shao, Q.-M.
\newblock \emph{Self-normalized processes: Limit theory and Statistical
  Applications}.
\newblock Springer Science \& Business Media, 2008.

\bibitem[Precup(2000)]{precup2000eligibility}
Precup, D.
\newblock Eligibility traces for off-policy policy evaluation.
\newblock \emph{Computer Science Department Faculty Publication Series}, pp.\
  ~80, 2000.

\bibitem[Puterman(2014)]{puterman2014markov}
Puterman, M.~L.
\newblock \emph{Markov decision processes: discrete stochastic dynamic
  programming}.
\newblock John Wiley \& Sons, 2014.

\bibitem[Robins et~al.(2008)Robins, Li, Tchetgen, van~der Vaart,
  et~al.]{robins2008higher}
Robins, J., Li, L., Tchetgen, E., van~der Vaart, A., et~al.
\newblock Higher order influence functions and minimax estimation of nonlinear
  functionals.
\newblock In \emph{Probability and statistics: essays in honor of David A.
  Freedman}, pp.\  335--421. Institute of Mathematical Statistics, 2008.

\bibitem[Robins et~al.(2017)Robins, Li, Mukherjee, Tchetgen, van~der Vaart,
  et~al.]{robins2017minimax}
Robins, J.~M., Li, L., Mukherjee, R., Tchetgen, E.~T., van~der Vaart, A.,
  et~al.
\newblock Minimax estimation of a functional on a structured high-dimensional
  model.
\newblock \emph{The Annals of Statistics}, 45\penalty0 (5):\penalty0
  1951--1987, 2017.

\bibitem[Sallab et~al.(2017)Sallab, Abdou, Perot, and Yogamani]{sallab2017deep}
Sallab, A.~E., Abdou, M., Perot, E., and Yogamani, S.
\newblock Deep reinforcement learning framework for autonomous driving.
\newblock \emph{Electronic Imaging}, 2017\penalty0 (19):\penalty0 70--76, 2017.

\bibitem[Shi \& Li(2021)Shi and Li]{shi2021testing}
Shi, C. and Li, L.
\newblock Testing mediation effects using logic of boolean matrices.
\newblock \emph{Journal of the American Statistical Association}, pp.\
  accepted, 2021.

\bibitem[Shi et~al.(2020{\natexlab{a}})Shi, Lu, and Song]{shi2020breaking}
Shi, C., Lu, W., and Song, R.
\newblock Breaking the curse of nonregularity with subagging---inference of the
  mean outcome under optimal treatment regimes.
\newblock \emph{Journal of Machine Learning Research}, 21\penalty0
  (176):\penalty0 1--67, 2020{\natexlab{a}}.

\bibitem[Shi et~al.(2020{\natexlab{b}})Shi, Wan, Song, Lu, and
  Leng]{shi2020does}
Shi, C., Wan, R., Song, R., Lu, W., and Leng, L.
\newblock Does the markov decision process fit the data: testing for the markov
  property in sequential decision making.
\newblock In \emph{International Conference on Machine Learning}, pp.\
  8807--8817. PMLR, 2020{\natexlab{b}}.

\bibitem[Shi et~al.(2020{\natexlab{c}})Shi, Zhang, Lu, and
  Song]{shi2020statistical}
Shi, C., Zhang, S., Lu, W., and Song, R.
\newblock Statistical inference of the value function for reinforcement
  learning in infinite horizon settings.
\newblock \emph{arXiv preprint arXiv:2001.04515}, 2020{\natexlab{c}}.

\bibitem[Su et~al.(2020)Su, Srinath, and Krishnamurthy]{su2020adaptive}
Su, Y., Srinath, P., and Krishnamurthy, A.
\newblock Adaptive estimator selection for off-policy evaluation.
\newblock In \emph{International Conference on Machine Learning}, pp.\
  9196--9205. PMLR, 2020.

\bibitem[Sutton \& Barto(2018)Sutton and Barto]{sutton2018reinforcement}
Sutton, R.~S. and Barto, A.~G.
\newblock \emph{Reinforcement learning: An introduction}.
\newblock MIT press, 2018.

\bibitem[Tang et~al.(2019)Tang, Feng, Li, Zhou, and Liu]{tang2019doubly}
Tang, Z., Feng, Y., Li, L., Zhou, D., and Liu, Q.
\newblock Doubly robust bias reduction in infinite horizon off-policy
  estimation.
\newblock In \emph{International Conference on Learning Representations}, 2019.

\bibitem[Thomas \& Brunskill(2016)Thomas and Brunskill]{thomas2016data}
Thomas, P. and Brunskill, E.
\newblock Data-efficient off-policy policy evaluation for reinforcement
  learning.
\newblock In \emph{International Conference on Machine Learning}, pp.\
  2139--2148, 2016.

\bibitem[Thomas et~al.(2015{\natexlab{a}})Thomas, Theocharous, and
  Ghavamzadeh]{thomas2015high}
Thomas, P., Theocharous, G., and Ghavamzadeh, M.
\newblock High confidence policy improvement.
\newblock In \emph{International Conference on Machine Learning}, pp.\
  2380--2388, 2015{\natexlab{a}}.

\bibitem[Thomas et~al.(2015{\natexlab{b}})Thomas, Theocharous, and
  Ghavamzadeh]{thomas2015high2}
Thomas, P.~S., Theocharous, G., and Ghavamzadeh, M.
\newblock High-confidence off-policy evaluation.
\newblock In \emph{Twenty-Ninth AAAI Conference on Artificial Intelligence},
  2015{\natexlab{b}}.

\bibitem[Tsiatis(2007)]{tsiatis2007semiparametric}
Tsiatis, A.
\newblock \emph{Semiparametric theory and missing data}.
\newblock Springer Science \& Business Media, 2007.

\bibitem[Uehara et~al.(2019)Uehara, Huang, and Jiang]{uehara2019minimax}
Uehara, M., Huang, J., and Jiang, N.
\newblock Minimax weight and q-function learning for off-policy evaluation.
\newblock \emph{arXiv preprint arXiv:1910.12809}, 2019.

\bibitem[Van Der~Laan \& Lendle(2014)Van Der~Laan and Lendle]{van2014online}
Van Der~Laan, M.~J. and Lendle, S.~D.
\newblock Online targeted learning.
\newblock 2014.

\bibitem[Van~der Vaart(2000)]{van2000asymptotic}
Van~der Vaart, A.~W.
\newblock \emph{Asymptotic statistics}, volume~3.
\newblock Cambridge university press, 2000.

\bibitem[Van Der~Vaart \& Wellner(1996)Van Der~Vaart and Wellner]{van1996weak}
Van Der~Vaart, A.~W. and Wellner, J.~A.
\newblock Weak convergence.
\newblock In \emph{Weak convergence and empirical processes}, pp.\  16--28.
  Springer, 1996.

\bibitem[Zhang et~al.(2020)Zhang, Janson, and Murphy]{zhang2020inference}
Zhang, K.~W., Janson, L., and Murphy, S.~A.
\newblock Inference for batched bandits.
\newblock \emph{arXiv preprint arXiv:2002.03217}, 2020.

\bibitem[Zou et~al.(2019)Zou, Xu, and Liang]{zou2019}
Zou, S., Xu, T., and Liang, Y.
\newblock Finite-sample analysis for sarsa with linear function approximation.
\newblock In \emph{Advances in Neural Information Processing Systems}, pp.\
  8665--8675, 2019.

\end{thebibliography}
\bibliographystyle{icml2021}

\appendix
\subsection{Third-Order Q-Estimator}
We detail the form of $\widehat{Q}_k^{(3)}$. According to the definition, we have $$\widehat{Q}^{(3)}_k=\frac{1}{|\mathbb{I}_k|T(|\mathbb{I}_k|T-1)}\sum_{\substack{i_1\in \mathbb{I}_k,0\le t_1<T\\i_2\in \mathbb{I}_k,0\le t_2<T \\ (i_1,t_1)\neq (i_2,t_2) }}\mathcal{D}_k^{(i_1,t_1)}\mathcal{D}_k^{(i_2,t_2)} \widehat{Q}_k.$$
For any state-action pair $(a,s)$, it follows that
\begin{eqnarray*}
	\widehat{Q}^{(3)}_k(a,s)=\frac{(1-\gamma)^{-1}}{|\mathbb{I}_k|T(|\mathbb{I}_k|T-1)}\sum_{\substack{i_1\in \mathbb{I}_k,0\le t_1<T\\i_2\in \mathbb{I}_k,0\le t_2<T \\ (i_1,t_1)\neq (i_2,t_2) }}\widehat{\tau}_{k}(A_{i_1,t_1},S_{i_1,t_1},a,s)
	\{R_{i_1,t_1} +\gamma
	\Mean_{a' \sim \pi(\cdot|S_{i_1,t_1+1})}
	\mathcal{D}_k^{(i_2,t_2)}\widehat{Q}_k(a',S_{i_1,t_1+1})\\- \mathcal{D}_k^{(i_2,t_2)}\widehat{Q}_k(A_{i_1,t_1},S_{i_1,t_1})\}+\frac{1}{|\mathbb{I}_k|T} \sum_{i_2\in \mathbb{I}_k,0\le t_2<T} \mathcal{D}_k^{(i_2,t_2)}\widehat{Q}_k(a,s).
\end{eqnarray*}
The right-hand-side is equal to
\begin{eqnarray*}
	\widehat{Q}_k(a,s)+\frac{(1-\gamma)^{-1}}{|\mathbb{I}_k|T}\sum_{\substack{i\in \mathbb{I}_k,0\le t<T}}\widehat{\tau}_{k}(A_{i,t},S_{i,t},a,s)
	\{R_{i,t} +\gamma
	\Mean_{a' \sim \pi(\cdot|S_{i,t+1})}
	\widehat{Q}_k(a',S_{i,t+1})- \widehat{Q}_k(A_{i,t},S_{i,t})\}\\
	+\frac{(1-\gamma)^{-2}}{|\mathbb{I}_k|T(|\mathbb{I}_k|T-1)}\sum_{\substack{i_1\in \mathbb{I}_k,0\le t_1<T\\i_2\in \mathbb{I}_k,0\le t_2<T \\ (i_1,t_1)\neq (i_2,t_2) }}\widehat{\tau}_{k}(A_{i_1,t_1},S_{i_1,t_1},a,s)
	\{\gamma\Mean_{a' \sim \pi(\cdot|S_{i,t+1})}\widehat{\tau}_{k}(A_{i_2,t_2},S_{i_2,t_2},a',S_{i_1,t_1+1})\\-\widehat{\tau}_{k}(A_{i_2,t_2},S_{i_2,t_2},A_{i_1,t_1},S_{i_1,t_1})+(1-\gamma) \widehat{\tau}_{k}(A_{i_2,t_2},S_{i_2,t_2},a,s)\}\\ \times \{R_{i_2,t_2}+\gamma
	\Mean_{a' \sim \pi(\cdot|S_{i_2,t_2+1})}
	\widehat{Q}_k(a',S_{i_2,t_2+1})- \widehat{Q}_k(A_{i_2,t_2},S_{i_2,t_2})\}.	
\end{eqnarray*}
\subsection{Definition of the $L_2$-norm Convergence}\label{sec:l2conv}
A sequence of variables $\{X_n\}_{n\ge 0}$ is said to converge in $L_2$-norm to $X$ if and only if $\Mean |X_n-X|^2\to 0$ as $n\to \infty$. 

A Q-estimator $\widehat{Q}$ is said to converge in $L_2$-norm to $Q^{\pi}$ at a rate of $(nT)^{-\alpha}$ if 
\begin{eqnarray*}
	\sqrt{\Mean_{(a,s)\sim p_{\infty}} \Mean |\widehat{Q}(a,s)-Q^{\pi}(a,s)|^2}=O\{(nT)^{-\alpha}\}.
\end{eqnarray*}
Similarly, a conditional density ratio estimator $\widehat{\tau}$ is said to converge in $L_2$-norm to $\tau^{\pi}$ at a rate of $(nT)^{-\alpha}$ if 
\begin{eqnarray*}
	\sqrt{\Mean_{(a,s)\sim p_{\infty}} \Mean_{(a^*,s^*)\sim p_{\infty}} \Mean |\widehat{\tau}(a,s,a^*,s^*)-\tau^{\pi}(a,s,a^*,s^*)|^2}=O\{(nT)^{-\alpha}\}.
\end{eqnarray*}
Finally, a marginalized density ratio estimator $\widehat{\omega}$ is said to converge in $L_2$-norm to $\omega^{\pi}$ at a rate of $(nT)^{-\alpha}$ if 
\begin{eqnarray*}
	\sqrt{\Mean_{(a,s)\sim p_{\infty}} \Mean |\widehat{\omega}(a,s)-\omega^{\pi}(a,s)|^2}=O\{(nT)^{-\alpha}\}.
\end{eqnarray*}


\subsection{Proof of Lemma \ref{lemma5}}
To simplify the presentation, in the proof we assume the data consist of independent tuples in Lemma \ref{lemma3}. With weakly dependent data, the aggregated bias will be upper bounded by the same order of magnitude (see the proof of Theorem \ref{thm_TR} for details). 

We first study the bias of the Q-estimator. 
We will prove a slightly stronger result, showing that
\begin{eqnarray}\label{eqn:lemma4assertion}
\Mean_{(a,s)\sim p_{\infty}}	|\Mean  \widehat{Q}_k^{(m)}(a,s)-Q^{\pi}(a,s)|^2=O\{(nT)^{-2\alpha_1-2(m-1)\alpha_2}\}. 
\end{eqnarray}
We prove this assertion by induction. Consider the case where $m=2$. By the doubly-robustness property, we have $Q^{\pi}(a,s)=\Mean [\widehat{Q}_k(a,s)+\widehat{\tau}_k(A_{i,t},S_{i,t},a,s)\{R_{i,t}+\Mean_{a'\sim \pi(\cdot|S_{i,t+1})}\widehat{Q}_k(a',S_{i,t+1})-\widehat{Q}_k(A_{i,t},S_{i,t})\}]$. It follows that
\begin{eqnarray}\label{eqn:lemma4eq1}
\begin{split}
\Mean \widehat{Q}^{(2)}_k(a,s)-Q^{\pi}(a,s)=\Mean \mathcal{D}^{(i,t)}_k \widehat{Q}_k(a,s)-Q^{\pi}(a,s)=\Mean \{\widehat{\tau}_k(A_{i,t},S_{i,t},a,s)-\tau^{\pi}(A_{i,t},S_{i,t},a,s)\}\\\times \{Q^{\pi}(A_{i,t},S_{i,t})-\gamma\Mean_{a'\sim \pi(\cdot|S_{i,t+1})}Q^{\pi}(a',S_{i,t+1}) +\gamma \Mean_{a'\sim \pi(\cdot|S_{i,t+1})}\widehat{Q}_k(a',S_{i,t+1})-\widehat{Q}_k(A_{i,t},S_{i,t})\}.
\end{split}	
\end{eqnarray}
By Cauchy-Schwarz inequality, $\Mean_{(a,s)\sim p_{\infty}}|\Mean \widehat{Q}^{(2)}(a,s)-Q^{\pi}(a,s)|^2$ is upper bounded by
\begin{eqnarray*}
	\Mean_{(a,s)\sim p_{\infty}} \Mean |\widehat{\tau}_k(A_{i,t},S_{i,t},a,s)-\tau^{\pi}(A_{i,t},S_{i,t},a,s)|^2\left\{2\Mean |\widehat{Q}_k(A_{i,t},S_{i,t})-Q^{\pi}(A_{i,t},S_{i,t})|^2\right.\\
	+\left.2\Mean_{a\sim \pi (\cdot|S_{i,t+1})}\Mean |\widehat{Q}_k(a,S_{i,t+1})-Q^{\pi}(a,S_{i,t+1})|^2\right\}.
\end{eqnarray*}
Under the convergence rate requirement, it is upper bounded by $\{(nT)^{-\alpha_1-\alpha_2}\}$. This proves the assertion with $m=2$. 

Suppose the assertion holds with $m=m_0\ge 2$. We aim to show it holds with $m=m_0+1$. Similar to \eqref{eqn:lemma4eq1}, since the data tuples are i.i.d., 
we have
\begin{eqnarray*}
	\begin{split}
		\Mean \widehat{Q}_k^{(m_0+1)}(a,s)-Q^{\pi}(a,s)=\Mean \mathcal{D}^{(i,t)}_k \Mean \widehat{Q}_k^{(m_0)}(a,s)-Q^{\pi}(a,s)=\Mean \{\widehat{\tau}_k(A_{i,t},S_{i,t},a,s)-\tau^{\pi}(A_{i,t},S_{i,t},a,s)\}\times \\ [Q^{\pi}(A_{i,t},S_{i,t})-\gamma\Mean_{a'\sim \pi(\cdot|S_{i,t+1})}Q^{\pi}(a',S_{i,t+1}) +\gamma \Mean_{a'\sim \pi(\cdot|S_{i,t+1})}\Mean \{\widehat{Q}_k^{(m_0)}(a',S_{i,t+1})|S_{i,t+1}\}-\Mean\{\widehat{Q}_k^{(m_0)}(A_{i,t},S_{i,t})|A_{i,t},S_{i,t}\}].
	\end{split}	
\end{eqnarray*}
By Cauchy-Schwarz inequality, $\Mean_{(a,s)\sim p_{\infty}}|\Mean \widehat{Q}^{(m_0+1)}(a,s)-Q^{\pi}(a,s)|^2$ is upper bounded by
\begin{eqnarray}\label{eqn:Qbias}
\begin{split}
\Mean_{(a,s)\sim p_{\infty}} \Mean |\widehat{\tau}_k(A_{i,t},S_{i,t},a,s)-\tau^{\pi}(A_{i,t},S_{i,t},a,s)|^2\left[2\Mean |\Mean \{\widehat{Q}_k^{(m_0)}(A_{i,t},S_{i,t})|A_{i,t},S_{i,t}\}-Q^{\pi}(A_{i,t},S_{i,t})|^2\right.\\
+\left.2\Mean_{a\sim \pi (\cdot|S_{i,t+1})}\Mean |\Mean \{\widehat{Q}_k^{(m_0)}(a,S_{i,t+1})|S_{i,t+1}\}-Q^{\pi}(a,S_{i,t+1})|^2\right].
\end{split}	
\end{eqnarray}
The above bound is of the order $O\{(nT)^{-2\alpha_1+2m_0 \alpha_2}\}$. The assertion is thus proven. 

We next consider the bias of the resulting value. Since $\eta_{\tiny{\textrm{TR}}}^{(m)}$ is a simple average of $\{\psi_{i,t}^{(m)}\}_{i,t}$, it suffices to provide an upper bound for $\psi_{i,t}^{(m)}$ for a given tuple $(i,t)\in \mathbb{I}_k$. We decompose $\widehat{Q}_k^{(m)}$ into the sum of the following two parts:
\begin{eqnarray*}
	{{|\mathbb{I}_k|T \choose (m-1)}}^{-1}\sum_{(i_l,t_l)=(i,t)~\textrm{for~some}~l} \mathcal{D}_k^{(i_1,t_1)}\cdots \mathcal{D}_k^{(i_{m-1},t_{m-1})} \widehat{Q}_k\\
	+	{{|\mathbb{I}_k|T \choose (m-1)}}^{-1}\sum_{(i_l,t_l)\neq (i,t)~\textrm{for~any}~l} \mathcal{D}_k^{(i_1,t_1)}\cdots \mathcal{D}_k^{(i_{m-1},t_{m-1})} \widehat{Q}_k.
\end{eqnarray*}
Since the functions $\widehat{Q}_k,\widehat{\tau}_k$ and the immediate rewards are uniformly bounded, the first term is upper bounded by
\begin{eqnarray*}
	c (m-1) {{|\mathbb{I}_k|T \choose (m-1)}}^{-1}{{|\mathbb{I}_k|T-1 \choose (m-2)}}=\frac{c (m-1)^2}{|\mathbb{I}_k| T}=O(n^{-1} T^{-1}),
\end{eqnarray*}
where $c$ denotes some positive constant. Similarly, we can show the second term can be well-approximated by 
\begin{eqnarray*}
	\widehat{Q}_{k,i,t}^{(m)}=(m-1){{|\mathbb{I}_k|T-1 \choose (m-2)}}^{-1}\sum_{(i_l,t_l)\neq (i,t)~\textrm{for~any}~l} \mathcal{D}_k^{(i_1,t_1)}\cdots \mathcal{D}_k^{(i_{m-1},t_{m-1})} \widehat{Q}_k,
\end{eqnarray*}
with the approximation error upper bounded by $O(n^{-1} T^{-1})$. 

Since $\psi_{i,t}^{(m)}$ is a linear function $\widehat{Q}^{(m)}$, we have $\max_{i,t} |\psi_{i,t}^{(m)}-\phi_{i,t}^{(m)}|=O(n^{-1} T^{-1})$ where $\phi_{i,t}^{(m)}$ is a version of $\psi_{i,t}^{(m)}$ with $\widehat{Q}^{(m)}$ replaced with $\widehat{Q}_{i,t}^{(m)}$. It suffices to show the bias $\max_{i,t} |\Mean \phi_{i,t}^{(m)}-\eta^{\pi}|$ converges at a rate of $(nT)^{-\alpha_1-(m-1)\alpha_2-\alpha_3}$. Since the tuples of indices $(i,t), (i_1,t_1),\cdots,(i_m,t_m)$ are different, the corresponding data observations are independent. This assertion can be proven in a similar manner as \eqref{eqn:lemma4assertion}. 

\subsection{Proof of Theorem \ref{thm_TR}}
For any $k$, let $r_1,r_2,r_3$ denote the rate of convergence of $\widehat{Q}_k$, $\widehat{\tau}_k$ and $\widehat{\omega}_k$, respectively. These rates of convergence will approach zero when the corresponding nuisance estimators are consistent. 

In Part 1, we prove a version Lemma \ref{lemma5} holds under the exponential $\beta$-mixing condition in (A1) as well. Specifically, the aggregated bias of the Q-estimator decays at a rate of $O(r_1r_2^{(m-1)})$, and the bias of the corresponding value estimator decays at a rate of $O(r_1r_2^{(m-1)}r_3)$. When one of the three estimated nuisance functions is consistent, the bias decays to zero. 

In Part 2, we show the variance of the value estimator decays to zero. By Chebyshev's inequality, this implies that our value estimator is consistent. The proof is thus completed.

\textbf{Part 1}. To simplify the proof, we assume $\mathbb{I}_k$ contains a single element $i$. The bias is given by
\begin{eqnarray*}
	{T\choose m-1}^{-1} \sum_{t_1<\cdots<t_{m-1}} (\Mean \mathcal{D}_{k}^{(i,t_1)}\cdots \mathcal{D}_{k}^{(i,t_{m-1})} \widehat{Q}_k-Q^{\pi}).
\end{eqnarray*}
We next apply Berbee's coupling lemma \citep[see e.g., Lemma 4.1 in][]{dedecker2002maximal} to bound the bias. Consider a given ordered tuple $(t_1,t_2,\cdots,t_{m-1})$. Following the discussion below  Lemma 4.1 in \citep{dedecker2002maximal}, we can construct i.i.d. data tuples $\{(S_{i,t_l}^0,A_{i,t_l}^0,R_{i,t_l}^0,S_{i,t_l+1}^0)\}_{1\le l\le m-1}$ such that the event
\begin{eqnarray*}
	(S_{i,t_l}^0,A_{i,t_l}^0,R_{i,t_l}^0,S_{i,t_l+1}^0)=(S_{i,t_l},A_{i,t_l},R_{i,t_l},S_{i,t_l+1}),\,\,\,\,\forall 1\le l\le m-1,
\end{eqnarray*}
holds with probability at least $1-\sum_{l=1}^{m-2}\beta(t_{l+1}-t_l-1)$ where $\beta(\cdot)$ denotes the $\beta$-mixing coefficients of $\{(S_{t},A_{t},R_{t})\}_{t\ge 0}$. This allows us to decompose each of the individual bias $|\Mean \mathcal{D}_{k}^{(i,t_1)}\cdots \mathcal{D}_{k}^{(i,t_{m-1})} \widehat{Q}_m-Q^{\pi}|$ into the following two terms
\begin{eqnarray*}
	|\Mean \mathcal{D}_{k}^{(i,t_1)}\cdots \mathcal{D}_{k}^{(i,t_{m-1})} \widehat{Q}_k-Q^{\pi}|\mathcal{I}\{(S_{i,t_l}^0,A_{i,t_l}^0,R_{i,t_l}^0,S_{i,t_l+1}^0)=(S_{i,t_l},A_{i,t_l},R_{i,t_l},S_{i,t_l+1}),\,\,\,\,\forall 1\le l \le m-1\}\\
	+	|\Mean \mathcal{D}_{k}^{(i,t_1)}\cdots \mathcal{D}_{k}^{(i,t_{m-1})} \widehat{Q}_k-Q^{\pi}|\mathcal{I}\{(S_{i,t_l}^0,A_{i,t_l}^0,R_{i,t_l}^0,S_{i,t_l+1}^0)\neq (S_{i,t_l},A_{i,t_l},R_{i,t_l},S_{i,t_l+1}),\,\,\,\,\exists 1\le l \le m-1\}.
\end{eqnarray*}
Based on Lemma \ref{lemma5}, the first term can be upper bounded by $O(T^{-\alpha_1-(m-1)\alpha_2})$. Under the boundedness property, the second term is upper bounded by $c \{\sum_{l=1}^{m-2}\beta(t_{l+1}-t_l-1)\}$ for some constant $c>0$. Averaging over all possible combinations of individual debiasing operators yields the following upper bound
\begin{eqnarray*}
	O(T^{-\alpha_*})+c{T\choose m-1}^{-1} \sum_{t_1<\cdots<t_{m-1}}\sum_{l=1}^{m-2}\beta(t_{l+1}-t_l-1).
\end{eqnarray*}
Under (A1), we have $\beta(t)=O(\rho^t)$ for some $0<\rho<1$ and any $t\ge 0$. The second term is upper bounded by $O(T^{-1})$. This yields the upper bound $O(T^{-\alpha_*})$ when $\mathbb{I}_k$ consists of a single element. In general, we can show the bias is upper bounded by $O\{(nT)^{-\alpha_*}\}$. Using similar arguments, we can show the bias of the value is upper bounded by $O\{(nT)^{-\alpha}\}$. This completes the proof for Part 1.

\textbf{Part 2}. For $1\le k\le \mathbb{K}$, let $\widehat{\eta}_{\tiny{\textrm{TR}},k}^{(m)}=(nT/\mathbb{K})^{-1} \sum_{i\in \mathbb{I}_k} \sum_{t=0}^{T-1} \psi_{i,t}^{(m)}$. By Cauchy-Schwarz inequality, it suffices to show the $\Var(\widehat{\eta}_{\tiny{\textrm{TR}},k}^{(m)})\to 0$ for each $k$.  Using similar arguments in the proof of Lemma \ref{lemma5}, we can show the difference  $(nT/\mathbb{K})^{-1} \sum_{i\in \mathbb{I}_k} \sum_{t=0}^{T-1} (\psi_{i,t}^{(m)}-\phi_{i,t}^{(m)})$ is upper bounded by $O(n^{-1} T^{-1})$. Consequently, it suffices to upper bound the variance of $\widehat{\eta}_{\tiny{\textrm{TR}},k,U}^{(m)}=(nT/\mathbb{K})^{-1} \sum_{i\in \mathbb{I}_k} \sum_{t=0}^{T-1} \phi_{i,t}^{(m)}$. 

A key observation is that, conditional on the estimators $\widehat{Q}_k$, $\widehat{\tau}_k$ and $\widehat{\omega}_k$, $\widehat{\eta}_{\tiny{\textrm{TR}},k,U}^{(m)}$ corresponds to an $m$-th order U-statistic. Under the given conditions, the kernel function associated with the U-statistic is uniformly bounded. We first consider the variance of $\widehat{\eta}_{\tiny{\textrm{TR}},k,U}^{(m)}$ conditional on the nuisance estimators. To simplify the proof, we similarly assume that $\mathbb{I}_k$ consists of a single trajectory, as in Part 1. By definition, the conditional variance is given by
\begin{eqnarray*}
	\left(\frac{m!}{T!}\right)^2 \sum_{\substack{\textrm{disjoint}~t_1,\cdots,t_m\\ \textrm{disjoint}~t_1',\cdots,t_m' }} \Cov\left( \Mean_{(a,s)\sim (\pi,\mathbb{G})}\mathcal{D}_{k}^{(i,t_1)}\cdots \mathcal{D}_{k}^{(i,t_{m-1})} \widehat{Q}_k(a,s)+\frac{1}{1-\gamma} \widehat{\omega}_k(A_{i,t_m},S_{i,t_m})\{R_{i,m}\right.\\ -\gamma \Mean_{a\sim \pi(\cdot|S_{i,m+1})}\mathcal{D}_{k}^{(i,1)}\cdots\mathcal{D}_{k}^{(i,m-1)} \widehat{Q}_k(a,S_{i,m+1})+\mathcal{D}_{k}^{(i,1)}\cdots\mathcal{D}_{k}^{(i,m-1)} \widehat{Q}_k(A_{i,t_m},S_{i,t_m})\},\\
	\Mean_{(a,s)\sim (\pi,\mathbb{G})}\mathcal{D}_{k}^{(i,t_1')}\cdots \mathcal{D}_{k}^{(i,t_{m-1}')} \widehat{Q}_k(a,s)+\frac{1}{1-\gamma} \widehat{\omega}_k(A_{i,t_m'},S_{i,t_m'})\{R_{i,t_m'}-\gamma\\ \left.\left.\times \Mean_{a\sim \pi(\cdot|S_{i,t_m'+1})}\mathcal{D}_{k}^{(i,1)}\cdots\mathcal{D}_{k}^{(i,m-1)} \widehat{Q}_k(a,S_{i,t_m'+1})+\mathcal{D}_{k}^{(i,t_1')}\cdots\mathcal{D}_{k}^{(i,t_m'-1)} \widehat{Q}_k(A_{i,t_m'},S_{i,t_m'})\}\right|\widehat{Q}_k,\widehat{\tau}_k,\widehat{\omega}_k\right),
\end{eqnarray*}
where $\Mean_{(a,s)\sim (\pi,\mathbb{G})}$ denotes the expectation by assuming $s\sim \mathbb{G}$ and $a\sim \pi(\cdot|s)$. Using similar arguments in Part 1, we can show that the above conditional variance decays to zero. In addition, $\Mean (\widehat{\eta}_{\tiny{\textrm{TR}},k,U}^{(m)}|\widehat{Q}_k,\widehat{\tau}_k,\widehat{\omega}_k)$ is to converge to $\eta^{\pi}$, when one of the nuisance estimator is consistent. Under the given conditions, $\widehat{\eta}_{\tiny{\textrm{TR}},k,U}^{(m)}$ is bounded. This further yields that $\Var\{\Mean (\widehat{\eta}_{\tiny{\textrm{TR}},k,U}^{(m)}|\widehat{Q}_k,\widehat{\tau}_k,\widehat{\omega}_k) \}\to 0$. Together with the fact that the conditional variance of $\widehat{\eta}_{\tiny{\textrm{TR}},k,U}^{(m)}$ decays to zero, the variance of $\widehat{\eta}_{\tiny{\textrm{TR}},k,U}^{(m)}$ decays to zero. The proof is thus completed. 

\subsection{Proof of Theorem \ref{thm_eff}}
In the proof of Theorem \ref{thm_TR}, we have shown that $\widehat{\eta}_{\tiny{\textrm{TR}},k}^{(m)}-\widehat{\eta}_{\tiny{\textrm{TR}},k,U}^{(m)}=O(n^{-1} T^{-1})$. This in turn implies that $\widehat{\eta}_{\tiny{\textrm{TR}}}^{(m)}-\widehat{\eta}_{\tiny{\textrm{TR}},U}^{(m)}=O(n^{-1} T^{-1})$ where $\widehat{\eta}_{\tiny{\textrm{TR}},U}^{(m)}$ is a simple average of $\{\widehat{\eta}_{\tiny{\textrm{TR}},k,U}^{(m)}\}_k$. It suffices to focus on $\widehat{\eta}_{\tiny{\textrm{TR}},U}^{(m)}$. 

The rest of the proof is divided into three parts. We first define $\widehat{\eta}_{\tiny{\textrm{TR}},U}^{(m),*}$ as a version of $\widehat{\eta}_{\tiny{\textrm{TR}},U}^{(m)}$ with the Q-, marginalized density ratio and conditional density ratio estimators replaced by their oracle values, and prove that  $\sqrt{nT}(\widehat{\eta}_{\tiny{\textrm{TR}},U}^{(m),*}-\eta^{\pi})\stackrel{d}{\to} N(0,\sigma^2)$. We next show that the difference $\widehat{\eta}_{\tiny{\textrm{TR}},U}^{(m),*}-\widehat{\eta}_{\tiny{\textrm{TR}},U}^{(m)}+\Mean \widehat{\eta}_{\tiny{\textrm{TR}},U}^{(m)}-\eta^{\pi}$
is $o_p\{(nT)^{-1/2}\}$. The assertion thus follows from an application of  Slutsky's theorem. Finally, in Part 3, we present the variance decomposition formula for $\Var(\widehat{\eta}_{\tiny{\textrm{TR}},U}^{(m),*})$. 

\textbf{Part 1:} A key observation is that, the oracle version of the estimator $\widehat{\eta}_{\tiny{\textrm{TR}},U}^{(m),*}-\eta^{\pi}$ corresponds to an $m$-th order U-statistic. The corresponding symmetric kernel function is given by
\begin{eqnarray*}
	h(\{(S_{i_j,t_j},A_{i_j,t_j},R_{i_j,t_j},S_{i_j,t_j+1})\}_{j=1}^m)=\frac{1}{m(1-\gamma)}\sum_{j=1}^m  \left[\Mean_{(a,s)\sim (\pi,\mathbb{G})}\prod_{l\neq j}\mathcal{D}^{(i_l,t_l)} Q^{\pi}(a,s) +\frac{1}{1-\gamma}\omega^{\pi}(A_{i_j,t_j},S_{i_j,t_j}) \right.\\\left.\times \left\{R_{i_j,t_j}+\gamma \Mean_{a\sim \pi(\cdot|S_{i_j,t_j+1})} \prod_{l\neq j} \mathcal{D}^{(i_l,t_l)} Q^{\pi}(a,S_{i_j,t_j+1}) 
	-\prod_{l\neq j} \mathcal{D}^{(i_l,t_l)} Q^{\pi}(A_{i_j,t_j},S_{i_j,t_j})\right\}\right]-\eta^{\pi}.
\end{eqnarray*}
Here, $\mathcal{D}^{(i_1,t_1)}$ denotes a version of $\mathcal{D}^{(i_1,t_1)}_k$ by replacing the estimator $\widehat{\tau}_k$ with the oracle value $\tau^{\pi}$. 
Under (A1) and the boundedness assumption in (A3), the conditions in Theorem 1 (c) of \citet{denker1983u} are satisfied. The asymptotic normality of 
$\widehat{\eta}_{\tiny{\textrm{TR}},U}^{(m),*}$ is thus proven. In addition, the asymptotic variance of $\sqrt{nT}(\widehat{\eta}_{\tiny{\textrm{TR}},U}^{(m),*}-\eta^{\pi})$ is given by $(nT)^{-1}m^2\Mean |\sum_{i,t}  h_1(S_{i,t},A_{i,t},R_{i,t},S_{i,t+1})|^2$ where
\begin{eqnarray*}
	h_1(s_1,a_1,r_1,s_1')=\Mean_{(s_2,a_2,r_2,s_2'),\cdots, (s_m,a_m,r_m,s_m')\stackrel{iid}{\sim} p_{\infty}} h(\{(s_j,a_j,r_j,s_j')\}_{j=1}^m).
\end{eqnarray*} 
Here, we use $p_{\infty}$ to denote the limiting distribution of the stochastic process $\{(S_t,A_t,R_t,S_{t+1})\}_{t\ge 0}$. 

Since the expectation of the temporal-difference error $r+\gamma \Mean_{a'\sim \pi(\cdot|s')} Q^{\pi}(a',s')-Q(a,s)$ is zero under the distribution $p_{\infty}$, the function $h_1(s_1,a_1,r_1,s_1')$ equals
\begin{eqnarray*}
	\frac{1}{m(1-\gamma)} \omega^{\pi}(a_1,s_1) \{r_1+\gamma \Mean_{a_1'\sim \pi(\cdot|s_1')} Q^{\pi}(a_1',s_1')-Q(a_1,s_1)\}.
\end{eqnarray*}
Consequently, the asymptotic variance $\sigma^2$ equals
\begin{eqnarray*}
	\frac{1}{nT(1-\gamma)^2} \Mean \left|\sum_{i,t}\omega^{\pi}(A_{i,t},S_{i,t})\{R_{i,t}+\gamma \Mean_{a'\sim \pi(\cdot|S_{i,t+1})} Q^{\pi}(a',S_{i,t+1})-Q^{\pi}(A_{i,t},S_{i,t}) \}\right|^2. 
\end{eqnarray*}
Under MA and CMIA, for any index $i$, the sequence of  temporal-difference errors $\{\varepsilon_{i,t}\}_{t\ge 0}=\{R_{i,t}+\gamma \Mean_{a'\sim \pi(\cdot|S_{i,t+1})} Q^{\pi}(a',S_{i,t+1})-Q^{\pi}(A_{i,t},S_{i,t})\}_{t\ge 0}$ forms a martingale difference sequence. As such, the elements in $\{\omega^{\pi}(A_{i,t},S_{i,t})\varepsilon_{i,t}\}_{t\ge 0}$ are pairwise uncorrelated. Consequently, 
\begin{eqnarray*}
	\sigma^2=\frac{1}{nT(1-\gamma)^2} \sum_{i,t}\Mean \left|\omega^{\pi}(A_{i,t},S_{i,t})\{R_{i,t}+\gamma \Mean_{a'\sim \pi(\cdot|S_{i,t+1})} Q^{\pi}(a',S_{i,t+1})-Q^{\pi}(A_{i,t},S_{i,t}) \}\right|^2,
\end{eqnarray*}
and is equal to \eqref{lower_bound}. This completes the proof for Part 1.

\textbf{Part 2:} For any $1\le k\le \mathbb{K}$, we similarly define $\widehat{\eta}_{\tiny{\textrm{TR}},k,U}^{(m),*}$ as the oracle version of $\widehat{\eta}_{\tiny{\textrm{TR}},k,U}^{(m)}$. 
In this Part, we focus on proving $\sqrt{nT}\{\widehat{\eta}_{\tiny{\textrm{TR}},k,U}^{(m)}-\widehat{\eta}_{\tiny{\textrm{TR}},k,U}^{(m),*}-\Mean (\widehat{\eta}_{\tiny{\textrm{TR}},k,U}^{(m)}|\widehat{Q}_k,\widehat{\tau}_k,\widehat{\omega}_k)+\eta^{\pi}\}=o_p(1)$. This in turn implies that $\sqrt{nT}(\widehat{\eta}_{\tiny{\textrm{TR}},k,U}^{(m)}-\widehat{\eta}_{\tiny{\textrm{TR}},k,U}^{(m),*}-\Mean \widehat{\eta}_{\tiny{\textrm{TR}},k,U}^{(m)}+\eta^{\pi})=o_p(1)$ and hence $\sqrt{nT}(\widehat{\eta}_{\tiny{\textrm{TR}},U}^{(m)}-\widehat{\eta}_{\tiny{\textrm{TR}},U}^{(m),*}-\Mean \widehat{\eta}_{\tiny{\textrm{TR}},U}^{(m)}+\eta^{\pi})=o_p\{(nT)^{-1/2}\}$.

We next show $\sqrt{nT}\{\widehat{\eta}_{\tiny{\textrm{TR}},k,U}^{(m)}-\widehat{\eta}_{\tiny{\textrm{TR}},k,U}^{(m),*}-\Mean (\widehat{\eta}_{\tiny{\textrm{TR}},k,U}^{(m)}|\widehat{Q}_k,\widehat{\tau}_k,\widehat{\omega}_k)+\eta^{\pi}\}=o_p(1)$. 
To simplify the proof, we assume $\mathbb{I}_k$ consists of a single element $i$. Note that $\widehat{\eta}_{\tiny{\textrm{TR}},k,U}^{(m)}-\widehat{\eta}_{\tiny{\textrm{TR}},k,U}^{(m),*}$ can be decomposed into the sum $\sum_{j=0}^m \widehat{\eta}_{j,k}$ where $\widehat{\eta}_{0,k}$ is the main effect term,  $\widehat{\eta}_{1,k}$ is the first-order linear term and $\widehat{\eta}_{j,k}$ is the high-order U-statistic for any $j\ge 2$. Specifically, 
\begin{eqnarray*}
	\widehat{\eta}_{0,k}=\Mean_{(a,s)\sim (\pi,\mathbb{G})} \{\widehat{Q}_k(a,s)-Q^{\pi}(a,s)\},
\end{eqnarray*}
corresponding to the difference between two plug-in estimators. Its conditional variance equals zero given $\widehat{Q}_k$ and we have $\widehat{\eta}_{0,k}=\Mean (\widehat{\eta}_{0,k}|\widehat{Q}_k)$. $(1-\gamma)\widehat{\eta}_{1,k}$ equals
\begin{eqnarray*}
	\begin{split}
		&\frac{1}{T}\sum_{t=0}^{T-1} \widehat{\omega}_k(A_{i,t},S_{i,t})[Q^{\pi}(A_{i,t},S_{i,t})-\widehat{Q}_k(A_{i,t},S_{i,t})-\gamma \Mean_{a\sim \pi(\cdot|S_{i,t+1})} \{Q^{\pi}(a,S_{i,t+1})-\widehat{Q}_k(a,S_{i,t+1})\} ]\\
		+&\frac{1}{T}\sum_{t=0}^{T-1} \Mean_{(a,s)\sim (\pi,\mathbb{G})} \widehat{\tau}_k(A_{i,t},S_{i,t},a,s)[Q^{\pi}(A_{i,t},S_{i,t})-\widehat{Q}_k(A_{i,t},S_{i,t})-\gamma \Mean_{a\sim \pi(\cdot|S_{i,t+1})} \{Q^{\pi}(a,S_{i,t+1})-\widehat{Q}_k(a,S_{i,t+1})\} ]\\
		+&\frac{1}{T}\sum_{t=0}^{T-1} \{\widehat{\omega}_k(A_{i,t},S_{i,t})+\Mean_{(a,s)\sim (\pi,\mathbb{G})}\widehat{\tau}_k(A_{i,t},S_{i,t},a,s)-2\omega^{\pi}(A_{i,t},S_{i,t})\}\varepsilon_{i,t}.
	\end{split}	
\end{eqnarray*}
Using similar arguments in the proof of Part 1, the conditional variance of the third line given $\widehat{\omega}_k$ and $\widehat{\tau}_k$ is equal to $T^{-1} \Mean \{\widehat{\omega}_k(A_{i,t},S_{i,t})+\Mean_{(a,s)\sim (\pi,\mathbb{G})}\widehat{\tau}_k(A_{i,t},S_{i,t},a,s)-2\omega^{\pi}(A_{i,t},S_{i,t})\}^2\varepsilon_{i,t}^2$. It is of the order $o_p(T^{-1})$ given that $\widehat{\omega}_k$ and $\widehat{\tau}_k$ coverages to $\omega^{\pi}$ and $\tau^{\pi}$, respectively. As such, we have 
\begin{eqnarray}\label{eqn:line1}
\begin{split}
&\frac{1}{T}\sum_{t=0}^{T-1} \{\widehat{\omega}_k(A_{i,t},S_{i,t})+\Mean_{(a,s)\sim (\pi,\mathbb{G})}\widehat{\tau}_k(A_{i,t},S_{i,t},a,s)-2\omega^{\pi}(A_{i,t},S_{i,t})\}\varepsilon_{i,t}\\
=&\Mean\left[\left.\frac{1}{T}\sum_{t=0}^{T-1} \{\widehat{\omega}_k(A_{i,t},S_{i,t})+\Mean_{(a,s)\sim (\pi,\mathbb{G})}\widehat{\tau}_k(A_{i,t},S_{i,t},a,s)-2\omega^{\pi}(A_{i,t},S_{i,t})\}\varepsilon_{i,t}\right|\widehat{\omega}_k,\widehat{\tau}_k\right]+o_p(T^{-1/2}).
\end{split}
\end{eqnarray}
As for the first line, similar to the proof of Theorem \ref{thm_TR}, we will apply Berbee's coupling lemma to bound its conditional variance. Specifically, following the discussion below Lemma 4.1 of \cite{dedecker2002maximal},  we can construct a sequence of data tuples $\{O_{i,t}^0=(S_{i,t_l}^0,A_{i,t_l}^0,R_{i,t_l}^0,S_{i,t_l+1}^0)\}_{1\le l\le m-1}$ such that
\begin{eqnarray}\nonumber
\frac{1}{T}\sum_{t=0}^{T-1} \widehat{\omega}_k(A_{i,t},S_{i,t})[Q^{\pi}(A_{i,t},S_{i,t})-\widehat{Q}_k(A_{i,t},S_{i,t})-\gamma \Mean_{a\sim \pi(\cdot|S_{i,t+1})} \{Q^{\pi}(a,S_{i,t+1})-\widehat{Q}_k(a,S_{i,t+1})\} ]\\\label{line2}
=\frac{1}{T}\sum_{t=0}^{T-1} \widehat{\omega}_k(A_{i,t}^0,S_{i,t}^0)[Q^{\pi}(A_{i,t}^0,S_{i,t}^0)-\widehat{Q}_k(A_{i,t}^0,S_{i,t}^0)-\gamma \Mean_{a\sim \pi(\cdot|S_{i,t+1}^0)} \{Q^{\pi}(a,S_{i,t+1}^0)-\widehat{Q}_k(a,S_{i,t+1}^0)\} ],
\end{eqnarray}
with probability at least $1-T\beta(q)/q$ such that the sequences $\{U_{i,2t}^0:i\ge 0\}$ and $\{U_{i,2t+1}^0:i\ge 0\}$ are i.i.d. where $U_i^0=(O_{i,tq}^0,O_{i,tq+1}^0,\cdots,O_{i,tq+q-1}^0)$. Due to the independence, the conditional variance of \eqref{line2} is upper bounded by $O_p(q^2 T^{-1-2\alpha_1})$, under Condition (A2). Take $q$ to be proportional to $\log T$, the probability $1-T\beta(q)/q$ will approach $1$, under Condition (A1). As such, the conditional variance of \eqref{line2} is $o_p(T^{-1})$ and we have
\begin{eqnarray*}
	\frac{1}{T}\sum_{t=0}^{T-1} \widehat{\omega}_k(A_{i,t}^0,S_{i,t}^0)[Q^{\pi}(A_{i,t}^0,S_{i,t}^0)-\widehat{Q}_k(A_{i,t}^0,S_{i,t}^0)-\gamma \Mean_{a\sim \pi(\cdot|S_{i,t+1}^0)} \{Q^{\pi}(a,S_{i,t+1}^0)-\widehat{Q}_k(a,S_{i,t+1}^0)\} ]\\
	=\Mean \left[\left.\frac{1}{T}\sum_{t=0}^{T-1} \widehat{\omega}_k(A_{i,t}^0,S_{i,t}^0)[Q^{\pi}(A_{i,t}^0,S_{i,t}^0)-\widehat{Q}_k(A_{i,t}^0,S_{i,t}^0)-\gamma \Mean_{a\sim \pi(\cdot|S_{i,t+1}^0)} \{Q^{\pi}(a,S_{i,t+1}^0)-\widehat{Q}_k(a,S_{i,t+1}^0)\} ]\right|\widehat{Q}_k,\widehat{\omega}_k\right]\\
	+o_p(T^{-1/2}).
\end{eqnarray*}
This in turn implies that
\begin{eqnarray}\label{eqn:line2}
\begin{split}
\frac{1}{T}\sum_{t=0}^{T-1} \widehat{\omega}_k(A_{i,t},S_{i,t})[Q^{\pi}(A_{i,t},S_{i,t})-\widehat{Q}_k(A_{i,t},S_{i,t})-\gamma \Mean_{a\sim \pi(\cdot|S_{i,t+1})} \{Q^{\pi}(a,S_{i,t+1})-\widehat{Q}_k(a,S_{i,t+1})\} ]\\
=\Mean \left[\left.\frac{1}{T}\sum_{t=0}^{T-1} \widehat{\omega}_k(A_{i,t},S_{i,t})[Q^{\pi}(A_{i,t},S_{i,t})-\widehat{Q}_k(A_{i,t},S_{i,t})-\gamma \Mean_{a\sim \pi(\cdot|S_{i,t+1})} \{Q^{\pi}(a,S_{i,t+1})-\widehat{Q}_k(a,S_{i,t+1})\} ]\right|\widehat{Q}_k,\widehat{\omega}_k\right]\\
+o_p(T^{-1/2}).
\end{split}
\end{eqnarray}
Using similar arguments, we can show the second line satisfies a similar relation as well. 
This together with \eqref{eqn:line1} and \eqref{eqn:line2} yields that $\widehat{\eta}_{1,k}=\Mean (\widehat{\eta}_{1,k}|\widehat{Q}_k,\widehat{\omega}_k,\widehat{\tau}_k)+o_p(T^{-1/2})$. 

$\widehat{\eta}_{2,k}$ equals $\{T(T-1)\}^{-1}\sum_{t_1\neq t_2} \widehat{\eta}_{2,t_1,t_2,k}$ where $(1-\gamma)^2\widehat{\eta}_{2,t_1,t_2,k}$ equals
\begin{eqnarray*}
	\gamma\Mean_{a\sim \pi(\cdot|S_{i,t_1+1})}[\{\widehat{\omega}_k(A_{i,t_1},S_{i,t_1})+\Mean_{(a,s)\sim (\pi,\mathbb{G})}\widehat{\tau}_k(A_{i,t_1},S_{i,t_1},a,s) \}\widehat{\tau}_k(A_{i,t_2},S_{i,t_2},a,S_{i,t_1+1})\\-2\omega^{\pi}(A_{i,t_1},S_{i,t_1})\tau^{\pi}(A_{i,t_2},S_{i,t_2},a,S_{i,t_1+1})]\varepsilon_{i_2,t_2}\\
	- [\{\widehat{\omega}_k(A_{i,t_1},S_{i,t_1})+\Mean_{(a,s)\sim (\pi,\mathbb{G})}\widehat{\tau}_k(A_{i,t_1},S_{i,t_1},a,s) \}\widehat{\tau}_k(A_{i,t_2},S_{i,t_2},A_{i,t_1},S_{i,t_1})\\-2\omega^{\pi}(A_{i,t_1},S_{i,t_1})\tau^{\pi}(A_{i,t_2},S_{i,t_2},A_{i,t_1},S_{i,t_1})]\varepsilon_{i_2,t_2}\\ +\{\widehat{\omega}_k(A_{i,t_1},S_{i,t_1})+\Mean_{(a,s)\sim (\pi,\mathbb{G})}\widehat{\tau}_k(A_{i,t_1},S_{i,t_1},a,s) \}\widehat{\tau}_k(A_{i,t_2},S_{i,t_2},A_{i,t_1},S_{i,t_1})\\
	\times \{Q^{\pi}(A_{i,t_2},S_{i,t_2})-\widehat{Q}_k(A_{i,t_2},S_{i,t_2})-\Mean_{a\sim \pi(\cdot|S_{i,t_2+1})} \{Q^{\pi}(a,S_{i,t_2+1})-\widehat{Q}_k(a,S_{i,t_2+1})\}\}\\
	-	\gamma\{\widehat{\omega}_k(A_{i,t_1},S_{i,t_1})+\Mean_{(a,s)\sim (\pi,\mathbb{G})}\widehat{\tau}_k(A_{i,t_1},S_{i,t_1},a,s) \}\Mean_{a\sim \pi(\cdot|S_{i,t_1+1})}\widehat{\tau}_k(A_{i,t_2},S_{i,t_2},a,S_{i,t_1+1})\\
	\times \{Q^{\pi}(A_{i,t_2},S_{i,t_2})-\widehat{Q}_k(A_{i,t_2},S_{i,t_2})-\Mean_{a\sim \pi(\cdot|S_{i,t_2+1})} \{Q^{\pi}(a,S_{i,t_2+1})-\widehat{Q}_k(a,S_{i,t_2+1})\}\}.
\end{eqnarray*}
Other high-order terms can be similarly derived. Using similar arguments in proving $\widehat{\eta}_{1,k}=\Mean (\widehat{\eta}_{1,k}|\widehat{Q}_k)+o_p(T^{-1/2})$, we can show  $\widehat{\eta}_{j,k}=\Mean (\widehat{\eta}_{j,k}|\widehat{Q}_k,\widehat{\omega}_k,\widehat{\tau}_k)+o_p(T^{-1/2})$ for any $j\ge 2$. This further implies that $\widehat{\eta}_{\tiny{\textrm{TR}},k,U}^{(m)}-\widehat{\eta}_{\tiny{\textrm{TR}},k,U}^{(m),*}-\Mean (\widehat{\eta}_{\tiny{\textrm{TR}},k,U}^{(m)}|\widehat{Q}_k,\widehat{\tau}_k,\widehat{\omega}_k)+\eta^{\pi}=o_p(T^{-1/2})$, since $\Mean \widehat{\eta}_{\tiny{\textrm{TR}},k,U}^{(m),*}=\eta^{\pi}$. More generally, when $\mathbb{I}_k$ consists of multiple trajectories, we can similarly show that $\widehat{\eta}_{\tiny{\textrm{TR}},k,U}^{(m)}-\widehat{\eta}_{\tiny{\textrm{TR}},k,U}^{(m),*}-\Mean (\widehat{\eta}_{\tiny{\textrm{TR}},k,U}^{(m)}|\widehat{Q}_k,\widehat{\tau}_k,\widehat{\omega}_k)+\eta^{\pi}=o_p(n^{-1/2}T^{-1/2})$. This completes the proof of Part 2. 

\textbf{Part 3:} Finally, we discuss the variance decomposition formula. Similar to Step 2, we can decompose $\widehat{\eta}_{\tiny{\textrm{TR}},U}^{(m),*}$ into the sum $\sum_{j=0}^m \widehat{\eta}^*_{j}$ where $\widehat{\eta}_0^*$ is the main effect $\eta^{\pi}=\Mean_{(a,s)\sim (\pi,\mathbb{G})} Q^{\pi}(a,s)$, $\widehat{\eta}_1^*$ is the first-order term  
\begin{eqnarray*}
	\frac{1}{nT(1-\gamma)}\sum_{i=1}^n\sum_{t=0}^{T-1}\omega^{\pi}(A_{i,t},S_{i,t})\{R_{i,t}+\gamma \Mean_{a\sim \pi(\cdot|S_{i,t+1})}Q^{\pi}(a,S_{i,t+1})-Q^{\pi}(A_{i,t},S_{i,t})\}.
\end{eqnarray*}
For any $j\ge 2$, $\widehat{\eta}_j^*$ corresponds to a degenerate U-statistic whose explicit form is given by
\begin{eqnarray*}
	{m\choose j} \frac{j!}{(nT)!}\sum_{\textrm{disjoint}~(i_1,t_1),\cdots,(i_j,t_j)} h_r(\{(S_{i_l,t_l},A_{i_l,t_l},R_{i_l,t_l},S_{i_l,t_{l+1}})\}_{l=1}^j),
\end{eqnarray*}
where 
\begin{eqnarray*}
	h_r(\{(s_l,a_l,r_l,s_l')\}_{l=1}^j)=\sum_{r=1}^j {j\choose r} (-1)^{j-r} \Mean_{(s_{l+1},a_{l+1},r_{l+1},s_{l+1}'),\cdots, (s_m,a_m,r_m,s_m')\stackrel{iid}{\sim} p_{\infty}} h(\{(s_j,a_j,r_j,s_j')\}_{j=1}^m),
\end{eqnarray*}
where the kernel $h$ is defined in Part 1. For instance, 
\begin{eqnarray*}
	\widehat{\eta}_2^*=\frac{1}{(1-\gamma)^2 nT(nT-1)} \sum_{(i_1,t_1)\neq (i_2,t_2)} \left[\omega^{\pi}(A_{i_1,t_1},S_{i_1,t_1})\{\gamma \Mean_{a'\sim \pi(\cdot|S_{i_1,t_1+1})} \tau^{\pi}(A_{i_2,t_2},S_{i_2,t_2},a',S_{i_1,t_1+1})\right.\\
	\left.-\tau^{\pi}(A_{i_2,t_2},S_{i_2,t_2},A_{i_1,t_1},S_{i_1,t_1})\}+(1-\gamma)\omega^{\pi}(A_{i_2,t_2},S_{i_2,t_2})\right]\epsilon_{i_2,t_2}.
\end{eqnarray*}
Other high-order terms can be similarly derived. 

\subsection{Proof of Theorem \ref{thm_adaptive}}
By Theorem \ref{thm_eff}, we have $\sqrt{nT}(\widehat{\eta}^{(m)}-\Mean \widehat{\eta}^{(m)})\stackrel{d}{\to} N(0,\sigma^2)$ for any $m$. Under the given conditions, using similar arguments in Part 1 of the proof of  Theorem \ref{thm_TR}, $\Mean \widehat{\eta}^{(m)}$ converges to $\eta^{\pi}$ at a rate of $o\{(nT)^{-1/2}\}$. This further implies that  $\sqrt{nT}(\widehat{\eta}^{(m)}-\eta^{\pi})\stackrel{d}{\to} N(0,\sigma^2)$. 

To prove the validity of our CI, it suffices to show the sampling variance estimator $(\widehat{\sigma}^{(m)})^2$ is consistent. The consistency can be proven using similar arguments in Part 2 of the proof of Theorem \ref{thm_eff}. We omit the details to save space. 

\section{More on the estimation of the nuisance functions}\label{sec_appendix_estimator}

\subsection{Fitted-Q evaluation}\label{secFQE}
We review the fitted-Q evaluation (FQE) algorithm proposed in \citet{le2019batch}, which is the subroutine we use to learn the Q-function. 
FQE is an iterative algorithm based on the  Bellman's equation: 
\begin{equation*}
Q(a, s) = \Mean_{a' \sim \pi(\cdot| s)} \left(R_t + \gamma Q(a'|S_{t+1})  | A_t = a, S_t = s \right). 
\end{equation*}
Based on this equation, we iteratively update the estimate by 
\begin{align*}
Q_m(a, s) 
= \argmin_{Q} 
\sum_{i'\in \mathbb{I}_k}\sum_{t < T}
\left\{
\gamma \Mean_{a' \sim \pi(\cdot| S_{i, t+1})} Q_{m-1}(a'|S_{i, t+1}) 
+   R_{i, t}  
- Q(A_{i, t}, S_{i, t})  \right\}^2, 
\end{align*}
for $m = 1, 2, \cdots$. 
The optimization problem can be solved with various supervised learning algorithms. 
We summarize FQE in Algorithm \ref{FQE}. 

\begin{algorithm}
	\caption{Fitted-Q evaluation}\label{FQE}
	\begin{algorithmic}
		\STATE \textbf{Input:} Data $\{S_{j,t},A_{j,t},R_{j,t},S_{j,t+1}\}_{j,t}$, policy $\pi$, function class $\mathcal{F}$, decay rate $\gamma$, number of iterations $M$
		
		\STATE Randomly pick $Q_0 \in \mathcal{F}$\;
		
		\FOR{$m = 1, \dots, M$}
		\STATE{Update target values 
			$Z_{j,t} = R_{j,t} + \gamma Q_{m-1}(S_{j,t+1}$, $\pi(S_{j,t+1}))$ for all $(j,t)$;}
		\STATE{Solve a regression problem to update the $Q$-function:\\
			$Q_m = \argmin_{Q \in \mathcal{F}} \frac{1}{n} \sum_{i=1}^n \{Q(S_{j,t}, A_{j,t}) - Z_{j,t}\}^2$
		}
		\ENDFOR
		\STATE \textbf{Output:} The estimated $Q$-function $Q_M(\cdot,\cdot)$
	\end{algorithmic}
\end{algorithm}	

\subsection{Learning the density ratio $\omega$}\label{secomega}

The estimation of the density ratio $\omega$ is based on the following key  observation. 
\begin{lemma}\label{lemma_omega}
	For any function $f$, we have $L(\omega, f) = 0$, where $L(\omega, f)$ is 
	\begin{equation}\label{eqn_omega1}
	\begin{split}
	\Mean_{a \sim \pi(\cdot|S_{i,t+1})} \{\omega(A_{i,t},S_{i,t})
	(\gamma f(a, S_{i,t+1})- f(A_{i,t},S_{i,t}) ) \} 
	+ (1-\gamma) \Mean_{S_0 \sim \mathbb{G}, a \sim \pi(\cdot|S_0)} f(a, S_0). 
	\end{split}
	\end{equation}
	Conversely, $\omega$ is the only function satisfying this condition. 
\end{lemma}
Therefore, as suggested in \citet{uehara2019minimax},  $\omega$ can be learned by solving the following mini-max problem
\begin{eqnarray}\label{optimize}
\argmin_{\omega\in \Omega} \sup_{f\in \mathcal{F}} L(\omega, f)^2, 
\end{eqnarray}
for some functional class $\Omega$ and $\mathcal{F}$. 
The expectation in \eqref{eqn_omega1} is approximated by the sample mean. To simplify the calculation, we can choose $\mathcal{F}$ to be a reproducing kernel Hilbert space (RKHS) , with which the inner maximization has a closed form solution, and then $\omega$ can be learned by solving the outer minimization via stochastic gradient descent. 
Let $\kappa(\cdot , \cdot; \cdot, \cdot)$ be the kernel function of the RKHS. 
Consider sampling a random minibatch $\{S_{i_g,t_g},A_{i_g,t_g},S_{i_g,t_g+1}:g\in \mathcal{M}\}$ from a data subset $\mathbb{I}_{k}$. 
We form the objective function $D(\omega)$ as  ${|\mathcal{M}| \choose 2}^{-1} \sum_{g_1,g_2\in \mathcal{M},g_1\neq g_2} D(\omega,g_1,g_2)$ where $D(\omega,g_1,g_2)$ is equal to
\begin{eqnarray*}
	2(1-\gamma) \omega(X_{i_{g_1},t_{g_1}})\Big\{\gamma\Mean_{\substack{a\sim \pi(\bullet|S_{i_{g_1},t_{g_1}+1})\\ s'\sim \mathbb{G},a'\sim \pi(\bullet|s') }} \kappa(S_{i_{g_1},t_{g_1}+1},a;s',a')-\Mean_{s'\sim \mathbb{G},a'\sim \pi(\bullet|s')}\kappa(X_{i_{g_1},t_{g_1}};s',a')\Big\}\\+\omega(X_{i_{g_1},t_{g_1}})\omega(X_{i_{g_2},t_{g_2}}) \Big\{ \gamma^2\Mean_{\substack{a_1\sim \pi(\bullet|S_{i_{g_1},t_{g_1}+1})\\a_2\sim \pi(\bullet|S_{i_{g_2},t_{g_2}+1}) }} \kappa(S_{i_{g_2},t_{g_2}+1},a_2;S_{i_{g_1},t_{g_1}+1},a_1) \\-2\gamma \Mean_{\substack{a\sim \pi(\bullet|S_{i_{g_1},t_{g_1}+1})}} \kappa(S_{i_{g_1},t_{g_1}+1},a;X_{i_{g_2},t_{g_2}})
	+\kappa(X_{i_2,t_2};X_{i_1,t_1})\Big\}\\+(1-\gamma)^2 \Mean_{\substack{s',s''\sim \mathbb{G}\\ a'\sim \pi(\bullet|s'),a''\sim \pi(\bullet|s'') }}\kappa(a',s';a'',s''),
\end{eqnarray*}
where  $X_{i,t}$ denotes the state-action pair $(A_{i,t},S_{i,t})$. 
Thus, in each step, we take a random minibatch from the observed data. Then we update the model parameter
\begin{eqnarray*}
	\theta\leftarrow \theta-\epsilon \Delta_{\theta} D(\omega_{\theta}/z_{\omega_{\theta}}), 
\end{eqnarray*}
where $z_{\omega_{\theta}}$ is a normalizing constant such that
\begin{eqnarray*}
	z_{\omega_{\theta}}=\frac{1}{|\mathcal{M}|} \sum_{g \in \mathcal{M} } \omega_{\theta}(A_{i_g,t_g},S_{i_g,t_g}).
\end{eqnarray*}
Note that $\omega$ satisfies $\Mean \omega(\pi,A_{t},S_{t})=1$. For a given $\widehat{\omega}_{k}$, we can further normalize the density ratio by $\widehat{\omega}_{k}(\bullet)=\widehat{\omega}_{k}(\bullet)/\{\sum_{j,t} \widehat{\omega}_{k}(A_{j,t},S_{j,t})/(nT) \}$. This yields the final estimates. 

\subsection{Learning the conditional sampling ratio $\tau$}\label{secomegastar}

Following the same analogy, our algorithm for estimating $\tau$ is  motivated by the following key observation. 
\begin{lemma}\label{lemma_omegastar1}
	For any two pairs $(i,t)$ and $(i',t')$ such that $O_{i,t}$ and $O_{i',t'}$ are independent, we have for any function $f$ that $\Mean  \Delta(\tau,f,\pi;i,t,i',t')=0$, where $\Delta(\tau,f,\pi;i,t,i',t')$ is

	\resizebox{0.9\linewidth}{!}{\begin{minipage}{\linewidth}
			\begin{equation*}
			\begin{split}
			\tau(S_{i',t'},A_{i',t'};A_{i,t},S_{i,t})\Big\{\gamma 
			\Mean_{a \sim \pi(\cdot|S_{i',t'+1})} f(S_{i',t'+1},a;A_{i,t},S_{i,t})\\-f(S_{i',t'},A_{i',t'};A_{i,t},S_{i,t})\Big\}+(1-\gamma)  f(A_{i,t},S_{i,t},A_{i,t},S_{i,t}).
			\end{split}
			\end{equation*}
	\end{minipage}}
	
	Conversely, $\tau$ is the only function satisfying this condition. 
\end{lemma}
Therefore, $\tau$ can be learned by solving the following mini-max problem
\begin{eqnarray}\label{optimize1}
\argmin_{\omega\in \Omega} \sup_{f\in \mathcal{F}} \left|\sum_{\substack{ (i,t)\neq (i',t')}} \Delta(\omega,f,\pi;i,t,i',t')\right|^2, 
\end{eqnarray}
for some functional class $\Omega$ and $\mathcal{F}$. 
The optimization for $\tau$ can be implemented in a similar way as that for $\omega$. 
Specifically, We set $\mathcal{F}$ to a unit ball of a reproducing kernel Hilbert space (RFHS), i.e., $\mathcal{F}=\{f\in \mathcal{H}:\|f\|_{\mathcal{H}}=1\}$, 
where 
\begin{eqnarray*}
	\mathcal{H}=\left\{f(\cdot)=\sum_{(i,t)\neq (i',t')} b_{i,t,i',t'} \kappa(X_{i',t'},X_{i,t};\cdot): b_{i,t,i',t'}\in \mathbb{R} \right\},
\end{eqnarray*}
for some positive definite kernel $\kappa(\cdot;\cdot)$, where $X_{i,t}$ is a shorthand for the state-action pair $(A_{i,t},S_{i,t})$. The optimization problem in \eqref{optimize1} is then reduced to
\begin{eqnarray*}
	\argmin_{\omega\in \Omega}  \sum_{ \substack{(i_1,t_1)\neq (i_1',t_1') \\ (i_2,t_2)\neq (i_2',t_2')} } D(\omega,\pi;i_1,t_1,i_1',t_1',i_2,t_2,i_2',t_2'),
\end{eqnarray*}

\noindent where $D(\omega,\pi;i_1,t_1,i_1',t_1',i_2,t_2,i_2',t_2')$ is given by
\begin{eqnarray*}
	\frac{\omega(X_{i_1',t_1'};X_{i_1,t_1})}{(1-\gamma)^{-1}}\Big\{\gamma\Mean_{a\sim \pi(\bullet|S_{i_1',t_1'+1})} \kappa(S_{i_1',t_1'+1},a,X_{i_1,t_1};X_{i_2,t_2},X_{i_2,t_2})-\kappa(X_{i_1',t_1'},X_{i_1,t_1};X_{i_2,t_2},X_{i_2,t_2})\Big\}\\+\frac{\omega(X_{i_2',t_2'};X_{i_2,t_2})}{(1-\gamma)^{-1}}\Big\{\gamma\Mean_{a\sim \pi(\bullet|S_{i_2',t_2'+1})} \kappa(S_{i_2',t_2'+1},a,X_{i_2,t_2};X_{i_1,t_1},X_{i_1,t_1})-\kappa(X_{i_2',t_2'},X_{i_2,t_2};X_{i_1,t_1},X_{i_1,t_1})\Big\}\\+\omega(X_{i_1',t_1'};X_{i_1,t_1})\omega(X_{i_2',t_2'};X_{i_2,t_2}) \Big\{ \gamma^2\Mean_{\substack{a_1\sim \pi(\bullet|S_{i_1',t_1'+1})\\a_2\sim \pi(\bullet|S_{i_2',t_2'+1}) }} \kappa(S_{i_2',t_2'+1},a,X_{i_2,t_2};S_{i_1',t_1'+1},a_1,X_{i_1,t_1}) \\-\gamma \Mean_{\substack{a_1\sim \pi(\bullet|S_{i_1',t_1'+1})}} \kappa(S_{i_1',t_1'+1},a,X_{i_1,t_1};X_{i_2',t_2'},X_{i_2,t_2})-\gamma \Mean_{\substack{a_2\sim \pi(\bullet|S_{i_2',t_2'+1})}} \kappa(S_{i_2',t_2'+1},a,X_{i_2,t_2};X_{i_1',t_1'},X_{i_1,t_1})\\
	+\kappa(X_{i_2',t_2'},X_{i_2,t_2};X_{i_1',t_1'},X_{i_1,t_1})\Big\}+(1-\gamma)^2 \kappa(X_{i_1,t_1},X_{i_1,t_1};X_{i_2,t_2},X_{i_2,t_2}).
\end{eqnarray*}
In our implementation, we set $\Omega$ to the class of neural networks. The detailed estimating procedure is given in Algorithm \ref{alg1}. 
\begin{algorithm}[t!]
	\caption{Estimation of the density ratio.}
	\label{alg1}
	\begin{algorithmic}
		\STATE \textbf{Input:} The data subset in $\mathcal{I}_{\ell}$.
		
		\STATE \textbf{Initial:} Initial the density ratio $\omega=\omega_{\beta}$ to be a neural network parameterized by $\beta$.
		\FOR{iteration $=1,2,\cdots$}
		\STATE{
			\STATE a. Randomly sample batches $\mathcal{M}$, $\mathcal{M}^*$ from the data transitions.
			\STATE b. {\textbf{Update}} the parameter $\beta$ by $$\beta\leftarrow \beta-\epsilon {|\mathcal{M}|\choose 2}^{-2}\sum_{\substack{(i_1,t_1),(i_1',t_1')\in \mathcal{M}\\ (i_1,t_1)\neq (i_1',t_1') }}\sum_{\substack{(i_2,t_2),(i_2',t_2')\in \mathcal{M}\\ (i_2,t_2)\neq (i_2',t_2') }} \nabla_{\beta} D(\frac{\omega_{\beta}}{z_{\omega_{\beta}}},\pi;i_1,t_1,i_1',t_1',i_2,t_2,i_2',t_2'),$$ 
			where $z_{\omega_{\beta}}$ is a normalization constant $$z_{\omega_{\beta}}(\cdot;A_{i,t},S_{i,t})=\frac{1}{|\mathcal{M}^*|} \sum_{(i',t') \in \mathcal{M}^* } \omega_{\beta}(X_{i',t'};X_{i,t}).$$ 
			
		}
		\ENDFOR
		\STATE \textbf{Output:} the density ratio $\omega_{\beta}$. 
		
	\end{algorithmic}
\end{algorithm}

\section{Additional numerical details}\label{sec:numerical_more_details}
In this section, we report more details of the simulation environments and the algorithm implementations. 

\subsection{More about the toy example}
The behaviour policy is chosen as a Bernoulli distribution with equal probabilities, and the target policy is chosen as follows: 
if the agent is at state A, then it takes action to transit to B or C with equal probabilities,  while if it is at state B or C, it takes action to transit to A with probability 1.0. 
The movement is uncertain: with probability 0.9 the transition will follow the action, and with 0.1 the agent will just stay where it is.  
The initial states are equally distributed over the three states. 
In Figure \ref{figure:toy_CI}, 
when the convergence rate of nuisance estimators is set as $(nT)^{-\alpha}$, 
to inject noises in the nuisance functions, 
we add a noise following  $\mathcal{N}(0, (0.2n^{-\alpha})^2)$ to $Q(s,a)$ when $Q$ is contaminated,
and add a noise following  $\mathcal{N}(0, (0.04 n^{-\alpha})^2)$ to the corrsponding density ratio when $\omega$ or $\tau$ is contaminated. 
In Figure \ref{figure:toy_triply}, 
to inject noises in the nuisance functions, 
we add a fixed noise following  $\mathcal{N}(0, 0.2^2)$ to $Q(s,a)$ when $Q$ is contaminated,
and add a fixed noise following  $\mathcal{N}(0, 0.04^2)$ to the corresponding density ratio when $\omega$ or $\tau$ is contaminated. 
The length of trajectories is fixed as $50$ for all settings. 

\subsection{More about the simulation settings} 
\subsubsection{The modified Cartpole environment}
Following \citet{uehara2019minimax}, we slightly modified the original Cartpole environment in \citet{brockman2016openai} to better fit the off-policy evaluation task. 
Specifically, we add small Gaussian noise with mean zero and standard deviation $0.02$ on the original deterministic transition dynamics, 
and define a new state-action-dependent reward as $(1 - (x ^ 2) / 11.52 - (\theta ^ 2) / 288)$,  where $x$ is the 
cart position and $\theta$ is the pole angle, to replace the original constant rewards. 

\subsubsection{The Diabetes environment}
We use the simulation environment about an mobile health application on diabetes control calibrated in \citet{shi2020does}. 
The state vector is $15$-dimensional and it contains the measurements of four hourly covariates and the hourly amounts of insulin injected in the past four hours, and the action space is discrete with $5$ levels on different amounts of insulin injection. 
The reward is a deterministic function of the glucose level, 
the state transition for the glucose is a linear function estimated from real data, 
and the noise for the glucose is set to have standard deviation $10$ in our experiment. 
The objective is to learn an optimal policy that maps patients' time-varying covariates into the amount of insulin injected to optimize patients' health status. 
More details can be found in \citet{shi2020does}. 

\subsubsection{Construction of the Behaviour and target policies}
For both environment, we first run deep-Q network to get a near-optimal $Q$-function $Q(s, a)$, and then apply softmax on its Q-value divided by an adjustable temperature $\tau$ to define the action probability of a behaviour policy as 
\begin{equation*}
\pi_b(a|s) \propto exp(\frac{Q(s, a)}{\tau})
\end{equation*}
For Cartpole, we model the $Q$-function as a dense neural network with $2$ hidden layers of dimension $256$, 
and set the optimizer as Adam with batch size $64$ and learning rate $0.01$. 
For Diabetes, we model the $Q$-function as a dense neural network with $2$ hidden layers of dimension $64$, 
and set the optimizer as Adam with batch size $128$ and learning rate $0.0001$.

\subsection{Implementation details}

For the Cartpole experiment, to implement our method, we set $\mathbb{K} = 2$ and sample $5\%$ of the total pairs in calculation of the incomplete U-statistics. 
To estimate the Q-function, we use random forests to model the Q-function, with the number of trees set as $1000$ and their max depth as $20$. 
To estimate $\omega$, we model it as a dense neural network with $5$ hidden layers of dimension $512$, connected via ReLu, and model the kernel $\mathrel{k}(\cdot, \cdot)$ as a Laplacian kernel with bandwidth chosen by the median heuristic. 
We optimize the problem via Adam with batch size $256$ and learning rate $0.001$. 
To estimate $\tau$, we model it as a dense neural network with $3$ hidden layers of dimension $512$, and optimize the problem via Adam with batch size $32$ and learning rate $0.0001$, with the other hyper-parameters the same with those of $\omega$. 

For the Diabetes experiment, to implement our method, we keep the other hyper-parameters the same with those for Cartpole, except that we sample $20\%$ of the total pairs in calculation of the incomplete U-statistics, adjust the  number of trees as $1000$ and their max depth as $50$, and 
adjust the learning rate for $\omega$ as $0.0001$ and the learning rate for $\tau$ as $0.00005$.  


To implement the IS-based CI construction methods, for simplicity, we directly use the true behaviour policies. 
The open-source code \footnote{\url{https://github.com/google-research/dice_rl}} is used to implement CoinDice. 
We use the default hyper-parameters, except for the following adjustments to get a better results for CoinDice. 
For CartPole, we set the learning rate as $0.005$, batch size as $32$, distribution regularizer as $0.05$, neural network regularizers as $1$, and set the neural networks as having one hidden layer of dimension $64$. 
For Diabetes, we adjust the distribution regularizer as $2.5$ and set the neural networks as having two hidden layers of dimension $256$. 
In our experiments, we find Coindice is sensitive to these hyper-parameters, and tuned intensively to report results with the best combination. 


\subsection{Computational complexity}\label{sec_computational_complexity}
In this section, we analyze the computational complexity for the proposed value estimator $\hat{\eta}^{(m)}_{TR}$. 
The construction of the CI is straightforward and has the same complexity. 
Let $N = nT$ and let the dimension of the action plus that of the state be $p$. 
There are four main dominating parts of the computation: 
the calculation of $\hat{Q}$, $\hat{\omega}$, and $\hat{\omega}^*$, and the construction of the final estimator. 
For simplicity, we assume the standard dense networks with feedforward pass and back-propagation are used for the first three parts, and 
let the maxmium latent layer width and the depth for all the neural networks be $w$ and $d$. 
For calculation of $\hat{Q}$, assume FQE converges in $M_1$ iterations, then according to the theory of neural networks, the complexity for the part is $O(NM_1w^dp)$. 
For calculation of $\hat{\omega}$ and $\hat{\omega}^*$, 
assume the training iterations of neural networks be $M_2$, 
then we have the complexity for these two part is $O(NM_2w^dp)$. 
For the last part, to calculate $\hat{\eta}^{(m)}_{TR}$, suppose we sample $M_3$ states from the reference distribution and use $M_4$ samples in the calculation of the incomplete U-statistics, the complexity is $O((M_3 + N)M_4)$. 
Putting the above results together, the total complexity for calculating $\hat{\eta}^{(m)}_{TR}$ and its CI is 
\begin{equation*}
O(nT(M_1 + M_2)w^dp + (M_3 + nT)M_4)
\end{equation*}
Note that the computation for the last part can be easily implemented in parallel, and for computing estimates of different order, the first three parts can be shared. 

\section{More on the CoinDice method}\label{sec:coindice}
We discuss why CoinDice would fail to achieve valid CI estimation in this section. As we have commented in the introduction, CoinDice uses the empirical likelihood approach  for interval estimation, assuming the data transactions are i.i.d. It is known that directly applying the empirical likelihood method without further adjustment will fail to handle weakly dependent data. 

To elaborate this, let us consider a simple example. Given a sequence of stationary random variables $\{Z_t\}_{1\le t\le n}$, we aim to construct a CI for its mean. The CI based on the empirical likelihood method is given as follows
\begin{eqnarray*}
	\{\Mean_{\mathbb{P}} Z: D_f(\mathbb{P}||\mathbb{P}_n)\le \rho/n\},
\end{eqnarray*}
for some $\rho>0$, where $\mathbb{P}_n$ denotes the empirical distribution of $\{Z_t\}_t$. 

Here, the choice of $\rho$ is essential to the validity of the resulting CI. When the observations $\{Z_t\}_t$ are i.i.d., one may set $\rho$ to $\prob(\chi_1^2\le \rho)=1-\alpha$ for a given significance $\alpha$. However, such a choice of $\rho$ would fail with weakly dependent observations. More specifically, $\rho$ shall be chosen such that
\begin{eqnarray*}
	\prob\left(\chi_1^2\le \frac{\rho \Var(Z_1)}{\Var(Z_1)+2\sum_{j=2}^{+\infty}\Cov(Z_1,Z_j)}\right)=1-\alpha,
\end{eqnarray*}
to ensure the validity of the resulting CI. See Theorem 5 and Theorem 11 of \citet{duchi2016statistics} for details. 

When the observations are weakly dependent, the factor $\Var(Z_1)/\{\Var(Z_1)+2\sum_{j=2}^{+\infty}\Cov(Z_1,Z_j)\}$ is not equal to one in general. Consequently, directly applying the empirical likelihood method by assuming the data are i.i.d. will result in an invalid CI. CoinDice estimates the value via the marginalized important-sampling estimator instead of the doubly-robust estimator. As such, the summands in their estimator are positively corrected. The corresponding factor is smaller than 1. Hence, applying CoinDice leads to a narrow but invalid CI. 

\end{document}